\documentclass[pdflatex,sn-mathphys-num]{sn-jnl}

\usepackage{graphicx}%
\usepackage{multirow}%
\usepackage{hyperref}
\usepackage{amsmath,amssymb,amsfonts}%
\usepackage{amsthm}%
\usepackage{mathrsfs}%
\usepackage[title]{appendix}%
\usepackage{xcolor}%
\usepackage{textcomp}%
\usepackage{manyfoot}%
\usepackage{booktabs}%
\usepackage{algorithm}%
\usepackage{algorithmicx}%
\usepackage{algpseudocode}%
\usepackage{listings}%
\usepackage{natbib}%

\theoremstyle{thmstyleone}%
\theoremstyle{thmstyletwo}%

\theoremstyle{thmstylethree}%

\begin{document}

\title[Article Title]{SolarBench: A global solar energy nowcasting benchmark}


\author*[1,2]{\fnm{Yuhao} \sur{Nie}}\email{nieyh@mit.edu}

\author[3]{\fnm{Stephen} \sur{Campbell}}

\author[4, 5]{\fnm{Quentin} \sur{Paletta}}
\author[6]{\fnm{Liwenbo} \sur{Zhang}}
\author[7]{\fnm{Tao} \sur{Jing}}
\author[8]{\fnm{Samer} \sur{Chaaraoui}}
\author[1]{\fnm{Jonathan} \sur{Giezendanner}}
\author[9]{\fnm{Andea} \sur{Scott}}
\author[10]{\fnm{Tao} \sur{Sun}}
\author[11]{\fnm{Cong} \sur{Feng}}
\author[12]{\fnm{Max} \sur{Aragon}}
\author[13]{\fnm{Jacques} \sur{Camier}}
\author[14]{\fnm{Adam} \sur{Jensen}}
\author[15]{\fnm{Florian} \sur{Kotthoff}}
\author[6]{\fnm{Yuexing} \sur{Yang}}
\author[6]{\fnm{Yang} \sur{Ming}}
\author[7]{\fnm{Mengying} \sur{Li}}
\author[8]{\fnm{Stefanie} \sur{Meilinger}}
\author[6]{\fnm{Yupeng} \sur{Wu}}
\author[9]{\fnm{Adam} \sur{Brandt}}
\author*[1,2,16]{\fnm{Sherrie} \sur{Wang}}\email{sherwang@mit.edu}

\affil[1]{\orgdiv{Institute for Data, Systems, and Society}, \orgname{Massachusetts Institute of Technology}, \orgaddress{\country{United States}}}

\affil[2]{\orgdiv{Laboratory for Information and Decision Systems}, \orgname{Massachusetts Institute of Technology}, \orgaddress{\country{United States}}}

\affil[3]{\orgdiv{Department of Electrical Engineering and Computer Science}, \orgname{Massachusetts Institute of Technology}, \orgaddress{\country{United States}}}

\affil[4]{\orgdiv{$\Phi$-lab, ESRIN}, \orgname{European Space Agency}, \orgaddress{\country{Italy}}}

\affil[5]{\orgdiv{Climate Office, ECSAT}, \orgname{European Space Agency}, \orgaddress{\country{United Kingdom}}}

\affil[6]{\orgdiv{Department of Architecture and Built Environment}, \orgname{University of Nottingham}, \orgaddress{\country{United Kingdom}}}

\affil[7]{\orgdiv{Department of Mechanical Engineering}, \orgname{Hong Kong Polytechnic University}, \orgaddress{\country{Hong Kong}}}

\affil[8]{\orgdiv{International Center for Sustainable Development}, \orgname{Hochschule Bonn-Rhein-Sieg University of Applied Sciences}, \orgaddress{\country{Germany}}}

\affil[9]{\orgdiv{Department of Energy Science and Engineering}, \orgname{Stanford University}, \orgaddress{\country{United States}}}

\affil[10]{\orgdiv{Department of Civil and Environmental Engineering}, \orgname{Stanford University}, \orgaddress{\country{United States}}}

\affil[11]{\orgdiv{Department of Mechanical Engineering}, \orgname{The University of Texas at Dallas}, \orgaddress{\country{United States}}}

\affil[12]{
\orgname{Mines Paris, Universit\'e PSL}, \orgaddress{\country{France}}}

\affil[13]{
\orgname{Independent Researcher}, \orgaddress{\country{United States}}}

\affil[14]{\orgdiv{Department of Civil and Mechanical Engineering}, \orgname{Technical University of Denmark}, \orgaddress{\country{Denmark}}}

\affil[15]{ \orgname{OFFIS - Institute for Information Technology}, \orgaddress{\country{Germany}}}

\affil[16]{\orgdiv{Department of Mechanical Engineering}, \orgname{Massachusetts Institute of Technology}, \orgaddress{\country{United States}}}

\abstract{As the share of solar power grows, nowcasting weather-driven solar variability becomes critical for reliable energy system operation. State-of-the-art approaches increasingly apply deep learning to sky camera and geostationary satellite observations, but fragmented datasets and inconsistent evaluation make it difficult to determine whether reported improvements generalize across climates, cloud regimes, and photovoltaic (PV) systems. Here we introduce SolarBench, an open global benchmark for image-based solar nowcasting. SolarBench harmonizes more than six million sky and satellite images from 11 diverse sites spanning a decade, together with irradiance or PV output and auxiliary atmospheric data. An accompanying toolbox supports reproducible data access, processing, model development, and evaluation. Using SolarBench, we benchmark representative models and reveal a gap between average forecasting accuracy and the ability to capture rapid solar fluctuations. We further quantify predictability across cloud regimes and demonstrate data-efficient adaptation to new PV systems. SolarBench provides an extensible foundation for fair comparison and methodological innovation in solar nowcasting.}


\maketitle

%
%

\section*{Main}\label{sec1}

Global electricity demand is growing rapidly, driven by rising consumption from industry, air conditioning, electric vehicles, and data centers \cite{IEA2026Electricity}. Meeting this demand while decarbonizing the energy system requires the large-scale integration of renewable generation \cite{chen2019pathways}. Solar photovoltaics (PV) are a key component of this transition because of their scalability, cost competitiveness, and low life-cycle emissions \cite{creutzig2017underestimated,Heath2012}. However, as solar deployment increases, weather-driven variability becomes an increasingly important operational challenge. Short-term forecasts over horizons of several minutes to 6 hours, commonly referred to as \textit{solar nowcasting} \cite{wang2026comprehensive}, provide a critical window for anticipating these fluctuations. Reliable nowcasts reduce uncertainty about solar generation before variability affects operational and economic decisions, supporting applications such as real-time grid balancing, energy trading, and flexible demand coordination \cite{venugopal2019short,gandhi2024value}.

A major challenge in solar nowcasting is resolving cloud dynamics \cite{barbieri2017very,xia2024accurate}. Solar generation can fluctuate sharply as clouds advect, form, and dissipate over PV systems \cite{le2020deep}. Under partly cloudy conditions, PV output can drop to near zero within minutes \cite{marquez2013intra}. Solar nowcasting approaches generally use recent irradiance or PV measurements, local meteorological observations, numerical weather predictions, or cloud imagery from ground-based sky cameras and geostationary satellites as inputs \cite{paletta2023advances,barhmi2024review}. Measurement-based approaches cannot observe clouds approaching from outside the site, while numerical weather prediction models generally operate at spatial and temporal resolutions too coarse to capture local, rapidly evolving cloud fields \cite{pedro2012assessment,sobri2018solar}. By contrast, sky cameras and geostationary satellites both provide observations of approaching clouds \cite{yang2018history}, with sky cameras resolving local conditions at higher frequency and satellites tracking cloud fields over broader regional scales at coarser update intervals \cite{chu2021intra} (Supplementary Fig. 1), offering complementary information on short-term solar variability. 
Advances in deep learning and computer vision have increasingly been adapted to solar nowcasting \cite{paletta2023advances}, with architectures evolving from convolutional and recurrent networks to multimodal and transformer-based approaches that combine imagery with recent solar measurements and auxiliary atmospheric data \cite{sun2019short,jiang2019deep,paletta2023omnivision,zhang2023advanced,nie2024skygpt}. These methods have shown strong performance in site-specific studies, and growing attention has focused on assessing their robustness across locations and adapting them to new PV systems \cite{sarmas2022transfer,nie2024sky}. 

Despite this progress, fragmented datasets and inconsistent evaluation make it difficult to determine whether reported performance gains generalize across climates, cloud regimes, and PV systems \cite{sun2019short}. Existing efforts have addressed different components of this challenge. Access to quality-controlled ground-based sky images and meteorological observations has improved in recent years \cite{nie2024open} through long-term observational archives \cite{andreas1981srrl,haeffelin2005sirta}, processed, ready-to-use datasets \cite{pedro2019comprehensive,nie2023skipp}, and software tools that facilitate data access and processing \cite{yang2018solardata,feng2019opensolar}. However, most ground-based datasets cover only one or a few sites and differ considerably in temporal resolution, record length, and available measurements \cite{nie2024open}. Geostationary satellite imagery provides another important input for solar nowcasting and is openly archived \cite{noaa2022goes,eumetsat_eumdac_guide,jma2022himawari8}, yet few resources provide processed datasets ready for use in nowcasting studies. SolarCube represents a recent effort to address this gap by integrating processed satellite imagery with solar radiation data across 19 regions and providing configurable dataset-generation tools and baseline benchmarks \cite{li2024solarcube}. Its scope, however, does not include ground-based sky imagery or integrated tools for model development and evaluation.

Benchmarking studies often rely on these heterogeneous resources and adopt study-specific forecast settings and evaluation protocols, hindering fair and consistent comparison \cite{paletta2021benchmarking,ruan2024use}. Moreover, irradiance remains the predominant prediction target, whereas system-level PV output is far less widely available because such data are often sensitive or proprietary \cite{nie2024open}. This scarcity leaves it unclear how readily methods developed for irradiance nowcasting can be adapted to newly deployed PV systems \cite{nie2024sky,palettaImprovingCrosssite2024}. To our knowledge, the field still lacks a multi-continental benchmark that combines broad cross-site coverage, harmonized multimodal data, and an end-to-end workflow for standardized model development and evaluation.

\begin{figure}
    \centering
    \includegraphics[width=1.\linewidth]{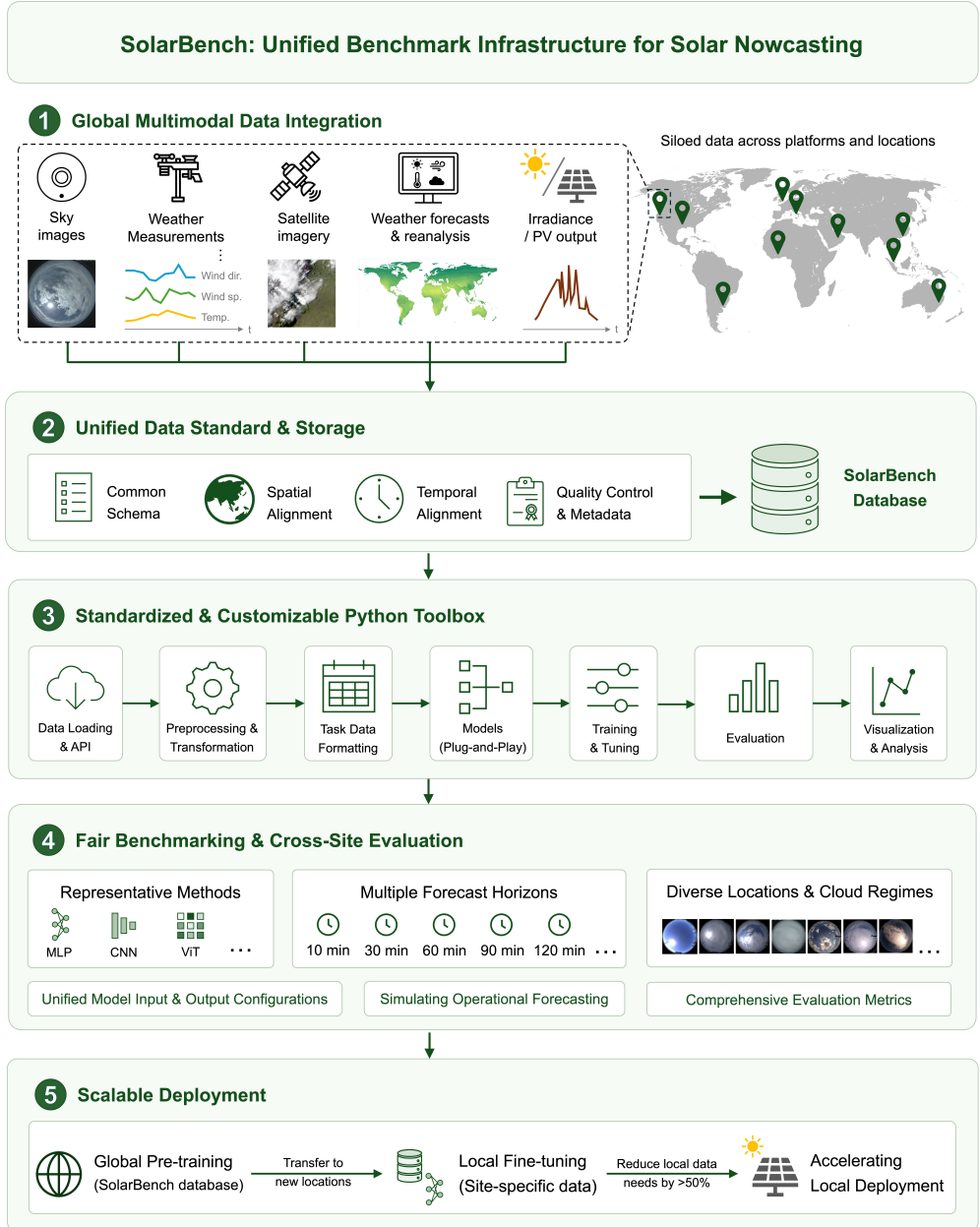}
    \caption{\textbf{Conceptual framework of SolarBench}. SolarBench provides a unified benchmark infrastructure for solar nowcasting that combines a curated multimodal database and an open machine learning toolbox for standardized data access, processing, model development, and evaluation. By combining a multi-continental harmonized database with reproducible benchmarking pipelines, SolarBench enables fair cross-site evaluation, reproducible research, and scalable development and cross-site adaptation of solar nowcasting systems.}
    \label{fig:conceptual_framework}
\end{figure}

Here we introduce SolarBench, an open global benchmark for image-based solar nowcasting (Fig.~\ref{fig:conceptual_framework}). SolarBench combines a curated multimodal database with a machine learning toolbox that standardizes the workflow from data access and processing to model development and evaluation. Using SolarBench, we show that state-of-the-art models improve average forecasting accuracy but fail to capture most rapid solar fluctuations. We further quantify how forecasting difficulty varies across sites and cloud regimes and demonstrate that geographically diverse pre-training enables data-efficient adaptation to new PV systems. These findings reveal limitations obscured by inconsistent and isolated site-specific evaluations and provide insights for the reliable deployment of solar nowcasting methods. SolarBench provides an extensible open foundation for researchers, forecasting providers, and energy stakeholders to advance methodological innovation, enable fair comparison, and support broader applications in solar nowcasting.

\section*{The SolarBench database and toolbox}\label{sec2}

\begin{figure}
    \centering
    \includegraphics[width=1.\linewidth]{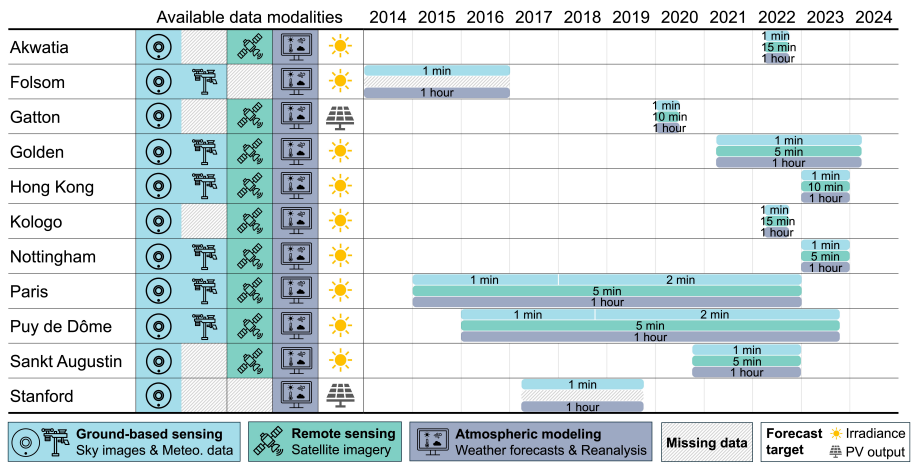}
    \caption{\textbf{Dataset characteristics}. 
\textbf{a}, Geographic distribution of the 11 SolarBench sites overlaid on long-term mean global downward surface solar radiation from ERA5 for 1980--2020. 
\textbf{b}, Short-term solar variability across sites, characterized by distributions of absolute 5-min changes in clear sky index ($|\Delta \mathrm{CSI}|$). Black diamonds indicate the root mean square of $|\Delta \mathrm{CSI}|$, defined as the variability index, and percentages indicate the fraction of 5-min intervals with $|\Delta \mathrm{CSI}| > 0.1$, used here to identify large short-term solar fluctuations.
\textbf{c}, Availability and temporal coverage of the three data categories across sites: ground-based observations, remote sensing data, and atmospheric model products, with irradiance or PV output indicated as the forecast target. Labels within the bars indicate the temporal resolution of the corresponding data.}
    \label{fig:dataset_characteristics}
\end{figure}

We collected data from 11 locations across five continents and seven countries to build the SolarBench database. The sites span seven K{\"o}ppen--Geiger climate subclasses across three of the five main climate groups---tropical, temperate, and continental \cite{beck2018present}---extend from approximately $28^\circ$S to $53^\circ$N, and cover a wide range of long-term solar-resource availability (Supplementary Table 2; Fig.~\ref{fig:dataset_characteristics}a). The sites also exhibit distinct patterns of short-term variability associated with cloud dynamics. To characterize this variability, we analyze absolute 5-min changes in the clear sky index (CSI), a normalized measure of irradiance or PV output relative to clear sky conditions that removes diurnal and seasonal trends in solar availability \cite{sun2019short}. Fig. \ref{fig:dataset_characteristics}b shows clear differences across sites in both the magnitude and frequency of short-term solar fluctuations. Although geographic coverage is not uniform, the database captures substantial diversity in climate, solar resource, and short-term variability, enabling systematic evaluation of model robustness across diverse forecasting environments.

For each site, SolarBench integrates three categories of data: ground-based observations, remote sensing data, and atmospheric model products (Fig. \ref{fig:dataset_characteristics}c). Ground-based observations include sky images, meteorological measurements, and solar irradiance or PV output data, generally available at 1--2 min resolution. Sky images and meteorological measurements provide high-frequency information on local atmospheric conditions, while irradiance or PV output serves as the prediction target for solar nowcasting tasks. Notably, SolarBench brings together system-level PV output from a rooftop installation \cite{nie2023skipp} and a multi-megawatt pilot-scale research facility with multiple array configurations \cite{xu2023gatton}, easing access to PV data that remain scarce in open solar nowcasting datasets and enabling evaluation across diverse PV deployment settings. Remote sensing data comprise geostationary satellite imagery from major operational platforms, including GOES \cite{noaa1994goesimager}, Himawari \cite{bessho2016introduction}, and Meteosat Second Generation (MSG) \cite{schmetz2002introduction}. These data are available at temporal resolutions of 5--15 min and capture regional cloud conditions surrounding each site. Atmospheric model products include hourly ECMWF Reanalysis v5 (ERA5) data \cite{c3s2023era5singlelevels} and operational weather forecasts from NOAA's Global Forecast System (GFS) \cite{noaa2004gfs}, providing large-scale atmospheric context across surface and upper-air levels. 

Across sites, temporal coverage ranges from about six months to eight years, with observations spanning 2014--2024 and capturing seasonal and interannual variability in solar conditions. In total, the processed database contains more than six million sky and satellite images, together with irradiance or PV output measurements and auxiliary atmospheric observations and model outputs, totaling approximately 500 GB of data. It is freely hosted on Hugging Face (\url{https://huggingface.co/solarbench}), a widely used open platform for sharing machine learning datasets and models. Additional details on site-specific data sources, collection protocols, and preprocessing procedures are provided in the Methods and Supplementary Information (SI).

The SolarBench toolbox is implemented in Python and provides standardized yet customizable modules for data access, preprocessing, model development, and evaluation. It implements operationally relevant forecasting task definitions, input--output configurations, train--test splits, and evaluation metrics to support consistent model comparison across sites and forecast settings (Methods). The database supports selective access by site and data modality, with documentation and demonstration examples provided to facilitate use. A contribution protocol supports the integration of new datasets, while the modular framework accommodates additional models, metrics, and benchmark tasks. The codebase is available on GitHub (\url{https://github.com/Solar4cast/solarbench}), and documentation is available at \url{https://solarbench.readthedocs.io}. These resources support reproducible and extensible use of the benchmark.

\section*{Enabling consistent and comprehensive benchmarking}\label{sec3}

We use solar irradiance nowcasting as a demonstration task for standardized benchmarking, with either global horizontal irradiance (GHI) or global tilted irradiance (GTI) as the prediction target depending on the measurements available at each site. Standardized nowcasting task definitions are described in the Methods. We benchmark representative methods spanning commonly used baselines and recent image-based deep learning architectures. These include SUNSET \cite{sun2019short}, a convolutional neural network (CNN) combining past sky image sequences with irradiance or PV measurement history; OmniVision \cite{paletta2023omnivision}, a multimodal CNN that additionally uses geostationary satellite imagery; and a Vision Transformer (ViT) adapted from \citet{zhang2023advanced} with temporal attention \cite{bertasius2021space} to jointly model sky image sequences and corresponding measurement histories. As baselines, we include smart persistence, which assumes a constant clear sky index over the forecasting horizon (hereafter simply referred to as persistence), and a lightweight multilayer perceptron (MLP) trained solely on historical measurements \cite{pedro2012assessment}. Detailed model architectures are described in the Methods, and model-specific implementation details, training procedures, and hyperparameter settings are provided in the SI.

We evaluate forecasting performance along two complementary dimensions: level prediction and ramp event prediction. Level prediction measures overall accuracy in estimating solar irradiance over the forecasting horizon, supporting baseline power prediction and proactive resource scheduling, and is quantified using forecast skill, defined as the improvement in root mean square error (RMSE) relative to persistence. Ramp event prediction assesses whether models capture rapid irradiance changes driven by cloud dynamics, which are important for anticipating sudden variability that can challenge grid stability. Existing studies predominantly emphasize level-based accuracy, while ramp prediction performance is less commonly reported and lacks standardized evaluation criteria \cite{chu2021intra}. To address this gap, SolarBench introduces an event-based ramp metric that evaluates whether predicted ramps match observed ramps in occurrence, direction, and magnitude within a prescribed relative error threshold (Methods).

\begin{figure}
    \centering
    \includegraphics[width=1.\linewidth]{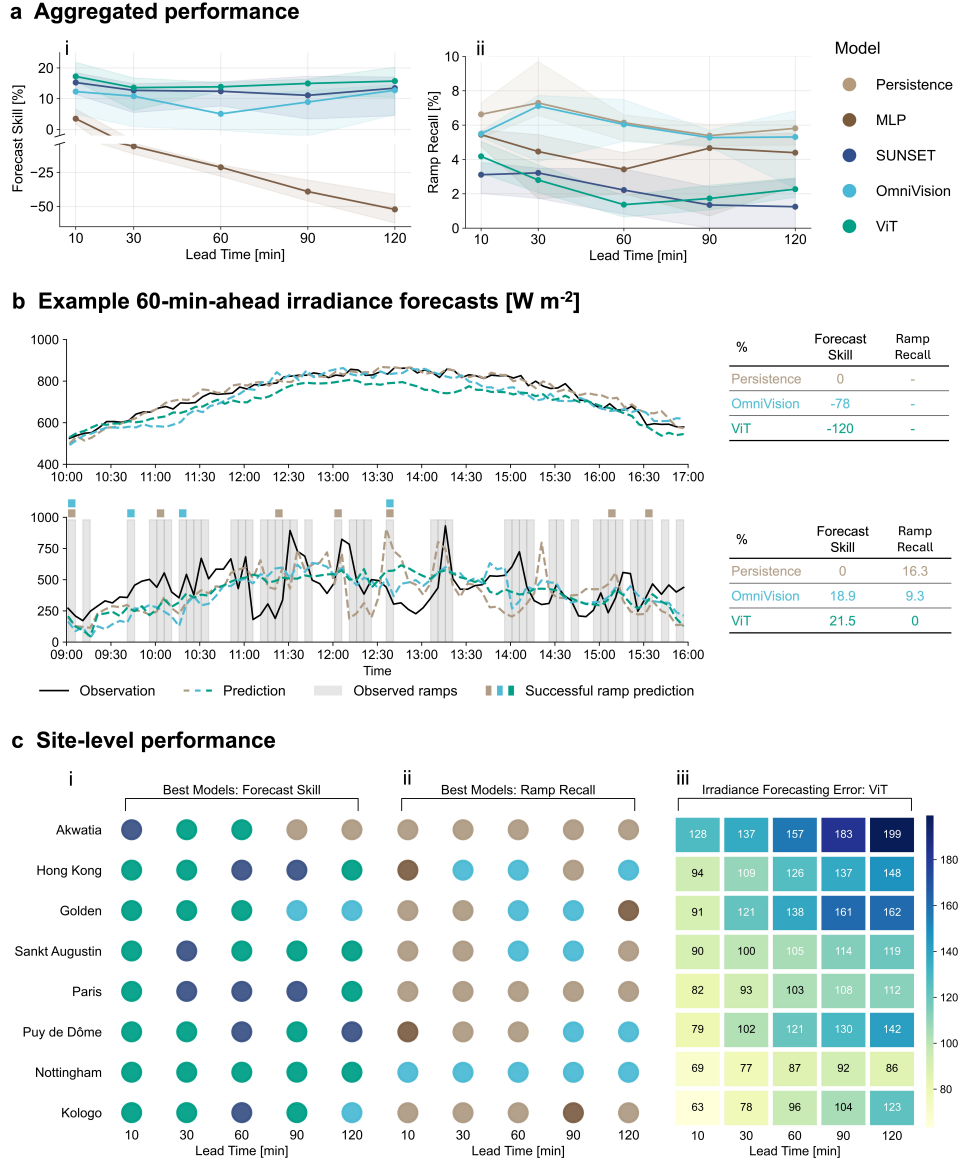}
    \caption{\textbf{Benchmarking of level and ramp prediction performance across lead times and locations.}
\textbf{a}, Aggregated forecasting performance. (i) Level prediction performance quantified by forecast skill (\%), defined as the improvement in RMSE related to persistence (i.e., persistence has a forecast skill of zero). Solid lines denote the median across locations, and shaded bands denote the interquartile range. (ii) Ramp recall, defined as the fraction of observed ramp events successfully captured within a 30\% relative error threshold, aggregated across sites in the same manner.
\textbf{b}, Example irradiance forecasts from representative models at a 60 min lead time under relatively stable and highly variable conditions. The stable case highlights the competitiveness of persistence under limited variability, whereas the highly variable case shows that image-based deep learning models achieve better level prediction accuracy but produce smoother forecasts that miss many rapid irradiance transitions, revealing a gap between aggregate accuracy and ramp event detection. Observed and successfully predicted ramp events are highlighted.
\textbf{c}, Site-level forecasting performance. (i) Best-performing model at each site and lead time for forecast skill and ramp recall. Each circle represents a site–lead time pair, and marker color denotes the best-performing model.
(ii), Irradiance forecasting RMSE of the ViT model across sites and lead times. Locations are sorted by RMSE at the 10 min lead time.}
    \label{fig:benchmark}
\end{figure}

Image-based information improves level prediction accuracy, although gains among advanced architectures remain modest. Across sites and lead times, image-based deep learning models, namely SUNSET, OmniVision, and ViT, improve over persistence by approximately 10–20\% in median forecast skill (Fig. \ref{fig:benchmark}a). Differences among these models remain relatively modest, generally within 3--6 percentage points. By contrast, the measurement-only MLP model exhibits strong performance degradation as lead time increases, highlighting the value of image-based information at longer horizons. In the short-term settings evaluated here, the tested multimodal architecture OmniVision does not consistently outperform sky image-based approaches, suggesting that implicitly learning radiative transfer from raw satellite images, as well as effective fusion of satellite and ground-based imagery, remain open modeling challenges.

Ramp event evaluation, however, reveals a pronounced accuracy-reliability gap. Despite improving aggregate forecast skill, image-based deep learning models do not consistently improve ramp recall and often capture fewer than 5\% of observed ramp events under a 30\% relative error threshold (Fig. \ref{fig:benchmark}a). The gap may partly reflect a mismatch between conventional training objectives, which are dominated by average level errors, and relatively infrequent but operationally important ramp events \cite{paletta2021benchmarking}. Evaluation based solely on aggregate error metrics can therefore obscure important forecasting limitations under highly variable cloud conditions. Example forecasts at a 60-min lead time illustrate this disconnect (Fig. \ref{fig:benchmark}b). Under relatively stable conditions, persistence remains highly competitive. Under highly variable conditions with frequent ramp events, deep learning models reduce aggregate error by producing smoother forecasts that tend toward intermediate irradiance levels, but this smoothing can miss many rapid transitions. Persistence produces larger level errors but carries recent irradiance fluctuations forward, leading to comparatively higher ramp recall. Its stronger ramp performance reflects preservation of recent variability rather than explicit anticipation of cloud evolution.

Model performance also varies substantially across locations, lead times, and evaluation metrics, with no single model consistently dominating under all conditions. For level prediction, the ViT is frequently the best-performing model, but SUNSET is favored at several lead times in Paris and Hong Kong, while OmniVision leads at longer horizons in Golden and Kologo. Ramp recall is more frequently dominated by persistence, although OmniVision performs best across lead times in Hong Kong and at selected horizons in Golden and Nottingham. The distribution of ViT RMSE further reveals heterogeneity in forecasting difficulty across sites and lead times (Fig. \ref{fig:benchmark}c), with median RMSE across locations increasing from 86 W m$^{-2}$ at 10 min to 133 W m$^{-2}$ at 120 min. These variations underscore the need for comprehensive benchmarking across diverse forecasting environments, horizons, and evaluation objectives, rather than drawing conclusions from a single-site or aggregate accuracy metric. Detailed site-level results for all models, lead times, and metrics are provided in the Supplementary Tables 6--10.

\section*{Revealing solar predictability across sites}\label{sec4}

Understanding what governs forecasting difficulty is critical to improving solar nowcasting systems. SolarBench enables predictability to be evaluated across geographically distributed sites and diverse atmospheric conditions. We first examine whether the variability index, a commonly used measure of fluctuations in solar resources \cite{stein2012variability}, also provides a useful indicator of forecasting difficulty. We find its relationship with forecasting error is weak and inconsistent across sites and lead times (Supplementary Fig. 6), indicating that aggregate variability alone does not adequately characterize predictability.

\begin{figure}
    \centering
    \includegraphics[width=1.\linewidth]{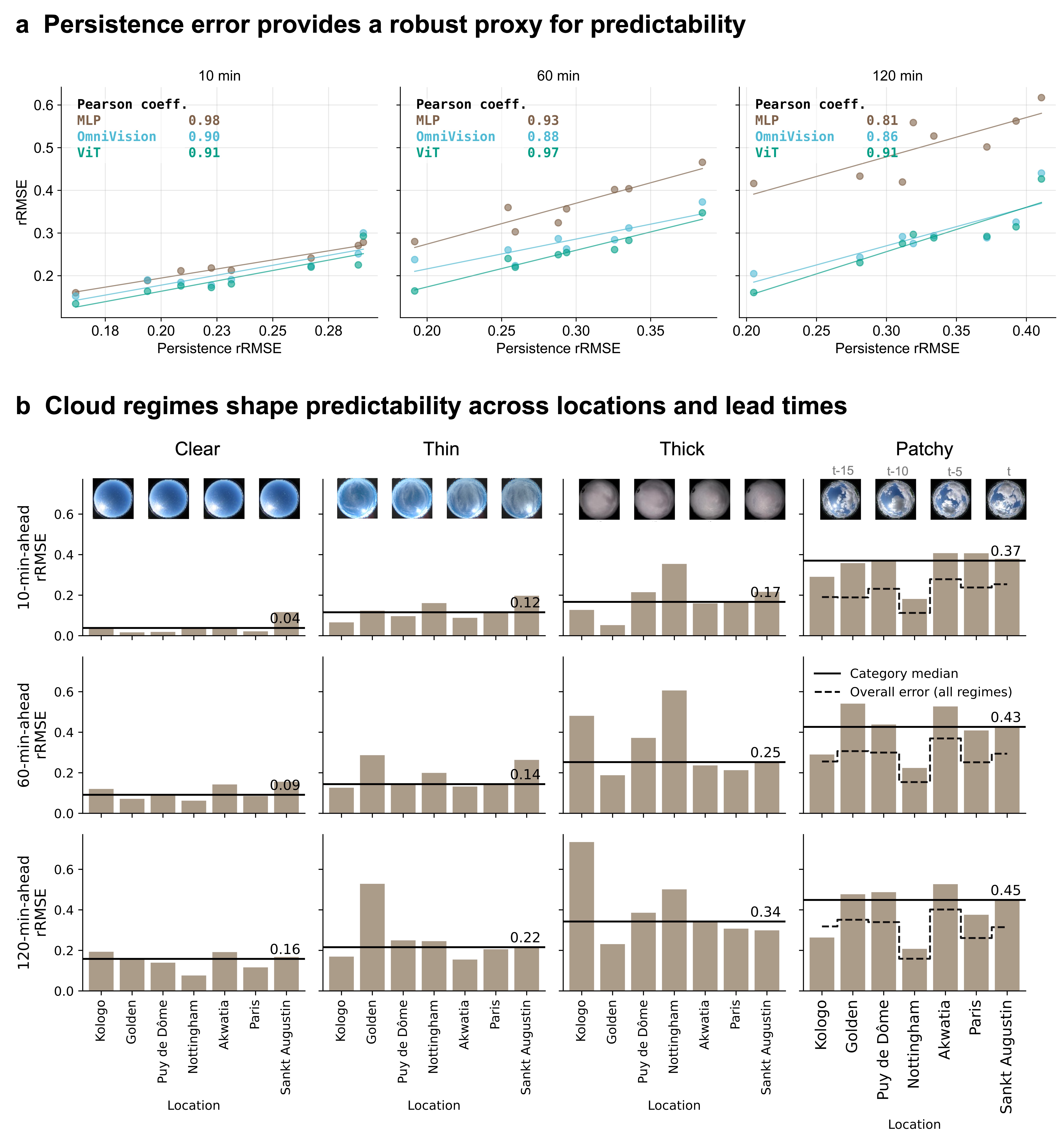}
    \caption{\textbf{Predictability across sites and cloud regimes}. 
\textbf{a}, Relationship between forecasting error and persistence-based predictability across sites for 10-, 60-, and 120-min lead times. Persistence rRMSE is used as a proxy for forecasting difficulty, with higher values indicating lower predictability. Results are shown for representative models (MLP, OmniVision, and ViT), with Pearson correlation coefficients reported for each lead time.
\textbf{b}, Persistence-based predictability across cloud regimes and lead times. Bar plots show persistence rRMSE across sites for Clear, Thin, Thick, and Patchy cloud regimes at 10-, 60-, and 120-min lead times. Solid horizontal lines indicate category median. Dashed lines in the Patchy panels denote the overall persistence error pattern across all cloud regimes, illustrating its close alignment with the cross-location error pattern under Patchy conditions. Representative sky image sequences for each cloud regime are shown above the first row.}
    \label{fig:model_predictability}
\end{figure}

Persistence-based error provides a more robust indicator of site-level forecasting difficulty. We quantify persistence error using relative root mean square error (rRMSE), which measures how well future irradiance can be inferred from the current atmospheric state. Higher persistence rRMSE indicates greater forecasting difficulty, or equivalently lower predictability. Because persistence error generally increases with lead time, comparisons are interpreted separately at each forecast horizon. At a 60-min lead time, for example, Nottingham shows the lowest persistence error (rRMSE=0.19), whereas Akwatia shows the highest (rRMSE=0.38), illustrating the range of site-level forecasting difficulty represented in SolarBench. Across models and forecast horizons, persistence rRMSE shows strong and consistent correlations with deep learning-based forecasting error, with Pearson correlation coefficients typically greater than 0.85 (Fig. \ref{fig:model_predictability}a). Persistence error therefore provides a simple and readily computed proxy for relative forecasting difficulty across diverse environments. 

Cloud regimes further explain systematic differences in predictability (Fig. \ref{fig:model_predictability}b). Here, we visually categorize cloud regimes as clear, thin, thick, or patchy based on the spatial structure and temporal evolution of cloud fields in historical sky image sequences (Methods). Clear sky and thin cloud conditions generally show low persistence errors, while thick cloud regimes correspond to intermediate error levels. Patchy cloud conditions generally produce the highest errors, indicating that spatially heterogeneous and rapidly evolving cloud fields pose the greatest forecasting challenge. These regime-dependent patterns vary across locations and forecast horizons. Nottingham is a notable outlier, with lower persistence errors under patchy clouds and higher errors under thick clouds. Persistence error also increases differently with lead time across sites, even within the same cloud regime. For example, under thick cloud conditions, Kologo shifts from below the regime median at 10 min to among the highest-error sites at 120 min, even exceeding its error under patchy clouds. 

Differences in overall predictability across sites closely mirror those observed under patchy cloud conditions. This alignment suggests that patchy cloud periods are particularly informative for diagnosing site-level forecasting difficulty. These findings motivate more targeted attribution of forecasting error under patchy cloud regimes. In particular, distinguishing errors associated with predominantly advective cloud motion from those driven by more complex cloud evolution could help determine whether forecast improvements require better characterization of cloud motion and spatial context, or richer information on cloud evolution and physical properties.

\section*{Lowering the local data barrier to deployment}
Limited local training data represent an important barrier to deploying site-specific solar nowcasting models at newly commissioned PV systems. SolarBench provides geographically distributed and consistently processed observations that support systematic analysis of local data requirements and cross-site knowledge transfer. Using the ViT, which showed the best overall level prediction performance in the benchmark above, as a representative image-based architecture, we examine how nowcasting performance scales with local data availability across sites with different levels of predictability. We then examine whether the knowledge learned through multi-site pre-training can be transferred to new PV systems with limited local observations. Details of the data scaling and transfer learning experiments are provided in the Methods.

\begin{figure}
    \centering
    \includegraphics[width=1.\linewidth]{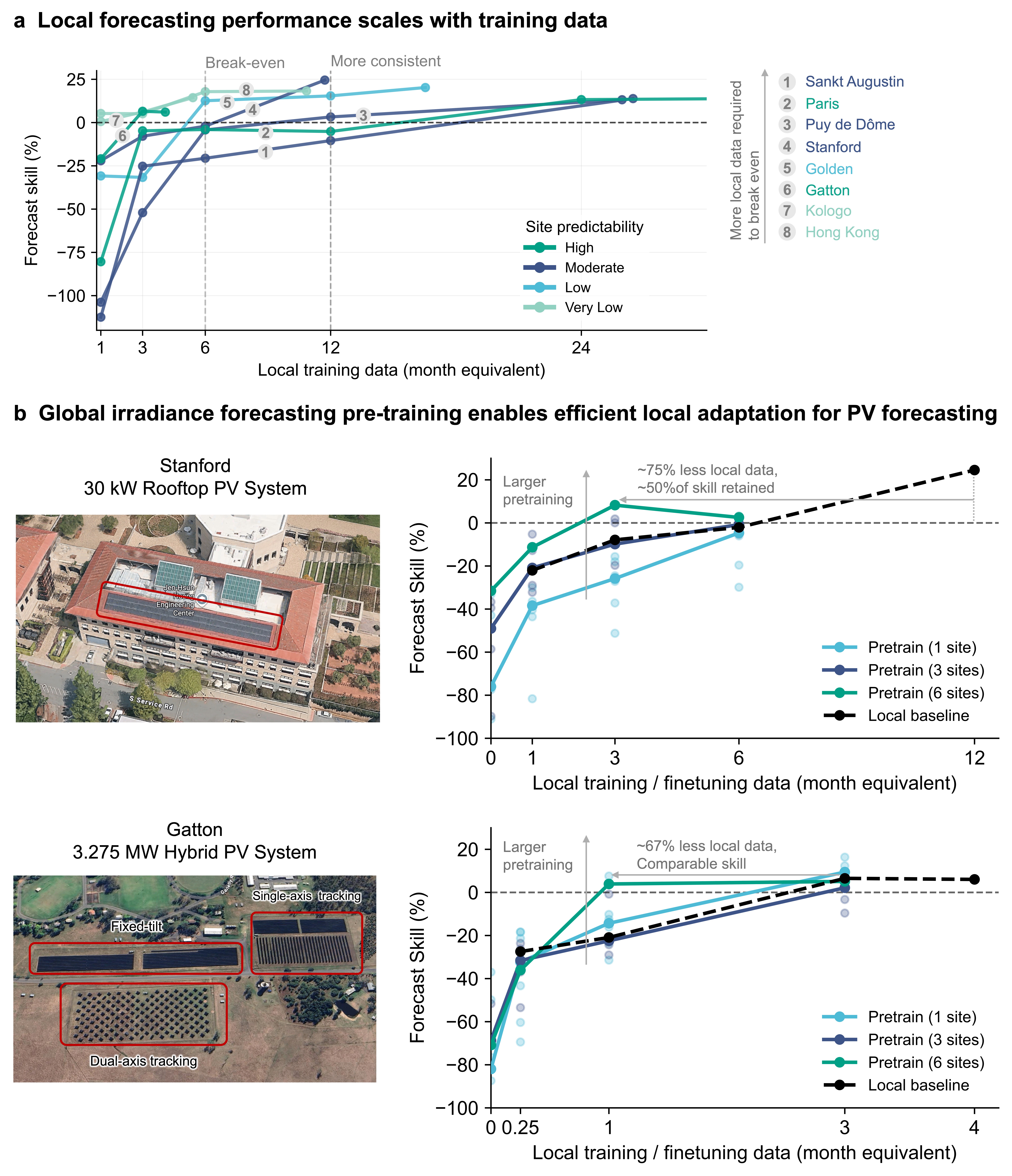}
    \caption{\textbf{Towards scalable deployment of solar forecasting systems.}
    \textbf{a}, Scaling of local forecasting performance with increasing training data across sites with different predictability levels. Deep learning models generally require approximately six months of local observations to achieve break-even performance relative to persistence, while more stable forecast skill typically emerges after $\sim$12 months of training data. Sites with higher predictability generally require more local data before deep learning models outperform persistence.
    \textbf{b}, Transfer learning from geographically distributed irradiance forecasting datasets to two real-world PV systems: a 30 kW rooftop PV installation at Stanford University, USA, and a 3.275 MW hybrid PV system at the Gatton Solar Research Facility at the University of Queensland, Australia. Pre-training on geographically distributed irradiance data followed by local fine-tuning substantially reduces local PV data requirements while maintaining competitive forecasting performance.}
    \label{fig:data_scaling_and_TL}
\end{figure}

We observe that forecast skill generally improves as additional training data become available before gradually saturating (Fig. \ref{fig:data_scaling_and_TL}a). Across sites, the model typically requires approximately six months of local observations to reach break-even performance relative to persistence, while more stable and consistent forecast skill generally emerges after one year of training data. Local data requirements vary with site predictability. At more predictable sites such as Sankt Augustin and Paris, where persistence is already a strong baseline, the model requires more data to outperform it; at less predictable sites such as Hong Kong and Kologo, benefits emerge with less local training data.

Geographically diverse pre-training substantially reduces this local data requirement. We first pre-train the ViT model on irradiance forecasting data from multiple SolarBench locations and subsequently fine-tune it for PV power forecasting using local PV observations. This setup reflects practical deployment conditions in which irradiance observations are more widely available, whereas PV generation data are often limited by ownership, privacy, and security constraints. We evaluate this framework on two contrasting PV systems: a 30 kW rooftop PV installation at Stanford University and a 3.275 MW hybrid PV system at the Gatton Solar Research Facility at the University of Queensland. 

Across both systems, multi-site pre-training combined with local fine-tuning enables data-efficient adaptation. Increasing the scale and diversity of the pre-training data consistently improves adaptation performance and reduces the amount of local data required by more than 50\%, while maintaining comparable forecast skill relative to models trained using the full local dataset (Fig. \ref{fig:data_scaling_and_TL}b). The benefits remain evident for the larger and more operationally complex Gatton system. These results show the potential of geographically distributed irradiance pre-training to lower the local data barrier across substantially different deployment settings and provide a scalable pathway for adapting forecasting models to newly deployed systems.

\section*{Conclusion} \label{sec5} 
SolarBench is an open global benchmark for image-based solar nowcasting that combines a curated multimodal database from 11 sites with a Python toolbox for standardized and reproducible model development and evaluation. The database integrates more than six million sky and satellite images, together with irradiance or PV output measurements and auxiliary atmospheric observations and model products, with records spanning up to a decade. Here, we demonstrate three applications of this resource: standardized multi-site benchmarking, predictability analysis across cloud regimes, and transfer learning for adapting forecasting models to operational PV systems.

The benchmarking results highlight the importance of a data-centric perspective for solar nowcasting. The largest improvement comes from adding visual observations that provide new sources of predictability beyond irradiance history, whereas differences among advanced deep learning architectures are comparatively modest, generally within 3–6 percentage points. At the same time, simply adding more modalities does not guarantee better performance: the tested satellite-sky image fusion model does not consistently outperform sky-image-based approaches in the nowcasting setting. Future progress will therefore require improvements at both the data and model layers, including more effective design of multimodal inputs and better fusion strategies for localized irradiance or PV output forecasting.

The benchmarking also reveals an important gap between average forecasting accuracy and operational reliability. Although image-based deep learning models achieve median forecast skill of 10--20\% relative to persistence, they do not consistently capture sudden irradiance ramps. Ramp recall remains low across tested models and lead times, with deep learning models often below 5\%. Models optimized primarily for error reduction may therefore still miss rapid ramps that affect operational decisions, such as reserve scheduling, battery storage dispatch, and demand response activation. This finding suggests that future solar nowcasting studies should move beyond aggregate error metrics by incorporating ramp-aware objectives and evaluation metrics that better reflect operational reliability.

SolarBench enables robust assessment of site-level predictability and provides insight into why forecasting difficulty varies across locations and atmospheric conditions. Persistence-based forecasting error emerges as a robust and easily computed proxy for relative predictability, while conventional variability metrics alone are less informative. Combined with globally available reanalysis data and clear sky models, this capability opens opportunities for large-scale predictability mapping and forecasting-aware solar asset planning. By linking predictability with cloud regimes, we further find that patchy, spatially heterogeneous cloud fields are a major source of forecasting difficulty. This connection between atmospheric dynamics and model performance points to future work on more targeted error attribution, including whether forecast improvements require better characterization of cloud motion, broader spatial context, or richer information on cloud evolution and physical properties.

For real-world deployment, limited local data remains a key constraint for newly deployed PV systems. Deep learning models often require about one year of local observations to achieve reliable performance. With SolarBench, we show that large-scale pre-training followed by lightweight local adaptation can reduce local data requirements by more than 50\% while maintaining competitive forecast performance. The benefits of geographically diverse pre-training further highlight the value of large-scale atmospheric observation networks and point toward transferable forecasting foundation models for solar energy applications. As low-cost sky cameras, satellite observations, and large-scale atmospheric datasets become increasingly accessible, this pathway could support scalable deployment of advanced solar nowcasting systems across diverse environments.

The current benchmark is not intended to be exhaustive. The models evaluated here are representative methods drawn from studies published up to 2023 with available code. The current site coverage, while multi-continental, remains nonuniform and is not intended to represent all global deployment environments. Future releases will expand toward broader geographic and climatic representation, particularly in underrepresented regions such as the Middle East and South America. SolarBench is designed to evolve through community contributions of additional datasets, models, metrics, and benchmark tasks. Future extensions could also complement accuracy-based benchmarking with forecast-value evaluation by combining SolarBench forecasts with downstream data, such as building loads, storage operation, electricity prices, or grid constraints, to assess how improvements in forecasting accuracy and reliability translate into operational and economic outcomes. By combining open data, reproducible workflows, and extensible evaluation protocols, SolarBench can help build a more comparable research ecosystem for solar nowcasting and support the development of reliable forecasting tools for large-scale solar energy integration.

\section*{Methods}\label{sec6}
\subsection*{Data collection, preprocessing, and standardization}
Data collection procedures differ across data types. Ground-based observations are compiled from published open datasets \cite{andreas1981srrl,haeffelin2005sirta,pedro2019comprehensive,baray2020cezeaux,nie2023skipp,xu2023gatton} and newly contributed datasets from collaborating research groups. Data collection details for the newly contributed datasets are provided in the SI. Geostationary satellite imagery, ERA5 reanalysis data, and GFS forecasts are obtained from publicly available archives, with satellite imagery sourced from GOES, Himawari, or MSG according to site location.

Raw data streams for each site are first processed and quality controlled before integration. Each record is associated with a timestamp, which is converted to a timezone-aware local time format to support site-specific analysis. Ground-based data are organized by temporally aligning sky images with available sensor measurements, including meteorological variables and irradiance or PV output. High-resolution sky images are down-sampled to 64 $\times$ 64 pixels, a resolution previously shown to reduce model training time while retaining acceptable forecasting performance \cite{sun2018solar}. For sites where image and sensor streams are collected at different temporal resolutions, sensor measurements are interpolated to image timestamps. Records associated with nighttime conditions, missing measurements, or physically implausible irradiance or PV output values are excluded. Full-disk or large-domain satellite imagery is cropped to a 256 km $\times$ 256 km context window centered on each site, and spectral bands with different native resolutions are resampled to 128 $\times$ 128 pixels. GFS and ERA5 variables are extracted from neighboring grid points around each site. Samples associated with timestamps containing missing values in required variables or spectral bands are excluded.

Processed data are partitioned into training and testing sets at the day level. For each site, roughly 5–10\% of available days are assigned to the test set, depending on dataset length. We select test days from across the full data collection period to capture seasonal and interannual variability, different sky conditions, and a range of short-term variability levels. Daily irradiance or PV output profiles were visually inspected during this process, with daily variability index values used as an additional guide. The remaining days are assigned to the training split, and the same train–test day assignments are applied consistently across all available data modalities.

The curated data are uploaded to Hugging Face and organized using a consistent structure to support efficient access through the SolarBench toolbox. Detailed information on data sources, available variables, preprocessing procedures, and train--test split dates is provided in the SI.

\subsection*{Standardized solar nowcasting tasks}
We define solar nowcasting tasks at short-term horizons relevant to operational power system decision-making, following established guidance for reproducible solar forecasting evaluation \cite{yang2019guideline}. For benchmarking efficiency, each task predicts solar irradiance or PV output at one of five representative lead times --- 10, 30, 60, 90, or 120 min --- rather than producing a fully time-resolved forecast trajectory. Although the benchmark evaluates individual lead times independently, the same framework can be extended to generate time-resolved forecast trajectories at finer temporal resolution by adjusting the task configuration in the SolarBench toolbox.

Model inputs are constructed by pairing imagery and non-imagery data within a 15-min historical input window at 5-min resolution, with the specific input modalities depending on the forecasting method. Each target is defined as the forward average over the subsequent 5-min interval at the specified lead time. Non-imagery scalar inputs are represented as backward averages over the corresponding 5-min intervals, whereas imagery inputs are aligned to the corresponding input timestamps. For methods that integrate sky and satellite imagery, differences in data cadence are handled by selecting the same number of satellite input steps as ground-based input steps, rather than enforcing an identical temporal window. As a result, the effective temporal span of satellite inputs can differ from the 15-min ground-based input window when satellite imagery has a coarser cadence.

These tasks are implemented in the SolarBench toolbox, which dynamically handles data loading, temporal alignment across modalities, sample construction, train--test splits, model training, and metric computation to enable standardized and reproducible benchmarking.

\subsection*{Benchmark models}
Across the analyses used to demonstrate SolarBench, models use different combinations of past sky and geostationary satellite image sequences together with irradiance or PV output history as inputs. Ground-based meteorological measurements, weather forecasts, and reanalysis data are included in SolarBench to support future studies of additional sources of predictability and multimodal forecasting methods.

As a common reference, we adopt a smart persistence model, hereafter simply referred to as persistence. The persistence model assumes that the clear sky index (CSI) remains constant over the forecast horizon and is widely used as a reference in solar nowcasting \cite{Chow2011,Chu2015,sun2019short}. The CSI is defined as
\begin{equation}\label{eq:csi}
    CSI_t = \frac{y_t}{y^{cs}_t}
\end{equation}
where $y_t$ denotes the observed target variable (solar irradiance or PV output) and $y^{cs}_t$ is the corresponding clear sky estimate. For irradiance targets, clear sky estimates are obtained from a clear sky irradiance model, such as the Ineichen–Perez model implemented in pvlib \cite{anderson2023pvlib}. For PV output targets, clear sky estimates are obtained either by combining clear sky irradiance with available PV system parameters to simulate clear sky power output, or by fitting a canonical geometric model to observed clear sky PV output data \cite{sun2019short}. These estimates capture deterministic solar geometry effects such as diurnal and seasonal variations. CSI values are clipped to the range [0.05, 1.5] to mitigate the impact of measurement noise and extreme outliers. Under the smart persistence assumption, the forecast at lead time $\tau$ is given by
\begin{equation}
\label{eq:persistence}
    \hat{y}_{t+\tau} = CSI_t \times y^{cs}_{t+\tau}
\end{equation} 

The MLP model is implemented as a lightweight feed-forward neural network trained only on historical irradiance or PV output measurements from the input window. The architecture follows \citet{pedro2012assessment}, with one hidden layer with 20 neurons and a ReLU activation function followed by a linear output layer.

SUNSET follows \citet{sun2019short} as a convolutional neural network that combines past sky image sequences with irradiance or PV output history. Sky images are processed using two convolutional blocks, each consisting of a convolutional layer, ReLU activation, batch normalization, and max pooling. The extracted image features are flattened and concatenated with the corresponding historical measurement inputs. The combined feature vector is then passed through two fully connected layers with dropout to generate the forecast.

OmniVision follows the multimodal spatiotemporal architecture proposed by \citet{paletta2023omnivision}. It jointly processes sequences of sky images, geostationary satellite images, and corresponding irradiance or PV output history. We use the first three available spectral bands from each satellite product, while additional bands retained in SolarBench are left for future studies. At each time step, the scalar irradiance or PV output measurement is repeated across the spatial dimensions and stacked with the three image channels, yielding a four-channel input for each modality. Separate convolutional and residual encoders extract spatial features from the sky and satellite inputs. The resulting feature sequences are processed independently by modality-specific temporal convolutional modules. The final temporal representations from the two modalities are concatenated and passed to a regression head to produce a single irradiance or PV output forecast.

The ViT model is adapted from \citet{zhang2023advanced} and extended with temporal attention inspired by TimeSFormer \cite{bertasius2021space}. For each input time step, the sky image is divided into non-overlapping patches and projected into patch embeddings, while the corresponding irradiance or PV output measurement is projected into a numerical token. Learnable spatial embeddings encode token positions within each frame, and learnable temporal embeddings distinguish the input time steps. The visual and numerical tokens from all frames are combined into a single sequence and processed by a transformer encoder with multi-head self-attention, allowing information to be integrated across spatial patches, measurements, and time. A global classification token aggregates information across space and time and the resulting representation is passed through a regression head to predict irradiance or PV output at the evaluated lead time.

To ensure fair comparison, all models are evaluated using the same train–test splits, input windows, lead times, forecasting targets, and evaluation metrics. Where available, implementations are based on publicly released code or code shared by the original authors, with adaptations made to support the standardized SolarBench task definitions and input formats. Model-specific implementation sources, adaptations, training procedures, and hyperparameters are provided in the SI.

\subsection*{Evaluation metrics}
Performance is evaluated using a comprehensive set of metrics. The primary metrics are described below, with additional metrics described in the SI.

For level prediction, we report RMSE together with two normalized metrics derived from it: forecast skill and rRMSE. RMSE quantifies absolute forecast error at each site in the units of the target variable, whereas forecast skill and rRMSE facilitate comparison across locations with different irradiance or PV output scales. RMSE is defined as
\begin{equation}
    RMSE = \sqrt{\frac{1}{N}\sum_{i=1}^{N} (y_i-\hat{y}_i)^2}
\end{equation}
where $N$ denotes the total number of test samples, and $y_i$ and $\hat{y}_i$ denote the observed and predicted target values, respectively. Forecast skill quantifies improvement over persistence (see Eq. \ref{eq:persistence}) and is defined as
\begin{equation}
   \mathrm{forecast\ skill} = \left(1 - \frac{RMSE}{RMSE_\mathrm{ref}}\right)\times 100\%
\end{equation}
where $RMSE_\mathrm{ref}$ denotes the RMSE of the persistence baseline. Positive values indicate improvement over persistence, whereas negative values indicate poorer performance. rRMSE normalizes RMSE by the root mean square of the observed target:
\begin{equation}
rRMSE =
\frac{RMSE}
{\sqrt{\frac{1}{N}\sum_{i=1}^{N}y_i^2}}
\end{equation}
Unlike the commonly used normalization by the mean observed target, this formulation accounts for differences in both the mean level and variability of the target.

For ramp event prediction, we evaluate how well each model captures short-term local variability in the target time series. We introduce an event-based ramp recall metric that combines detection of rapid changes with constraints on their direction and magnitude. An observed ramp is counted as correctly predicted only when the predicted change exceeds the threshold, has the same direction as the observed change, and reproduces its magnitude within a prescribed relative error tolerance. 

Observed and predicted ramp magnitudes are defined as the difference between consecutive values of the target variable
\begin{equation}
    r_i = y_i - y_{i-1}, \quad \hat{r}_i = \hat{y}_i - \hat{y}_{i-1}
\end{equation}
where consecutive indices $i$ and $i-1$ correspond to a 5-min interval in this study.

Ramp events are commonly defined by applying a threshold to ramp magnitude \cite{nouri2024ramp}. Rather than applying this threshold directly to $r_i$, we use the CSI (see Eq. \ref{eq:csi}) to remove deterministic solar geometry–induced trends and isolate cloud-driven variability. The observed and predicted CSI-based ramp magnitudes are defined as
\begin{equation}
    r^{CSI}_i = CSI_i - CSI_{i-1}, \quad \hat{r}^{CSI}_i = \widehat{CSI}_i - \widehat{CSI}_{i-1}
\end{equation}
To avoid numerical instability in CSI under low solar elevation, samples with solar elevation angle below $10^\circ$ are excluded from the ramp analysis. An observed ramp event is identified when the absolute CSI-based ramp magnitude exceeds a predefined threshold $\epsilon = 0.1$. Depending on the location, such events account for around 10--24\% of the test samples.

The set of observed ramp events is therefore constructed as
\begin{equation}
    \mathcal{E} = \left\{ i \;\middle|\; |r^{CSI}_i| > \epsilon \right\}
\end{equation}
where $M=|\mathcal{E}|$ is the number of observed ramp events.

For each observed ramp event, the relative ramp-magnitude error is defined as
\begin{equation}
RE_i
=
\frac{\left|\hat{r}_i-r_i\right|}{\left|r_i\right|} \times 100\%
\end{equation}

Ramp event recall, hereafter referred to as ramp recall, is defined as the fraction of observed ramp events correctly predicted:
\begin{equation}
\text{Ramp recall}
=
\frac{TP}{M} \times 100\%
=
\frac{TP_{\mathrm{up}}+TP_{\mathrm{down}}}{M} \times 100\%
\end{equation}
Here, $TP$ is the number of observed ramp events satisfying the ramp threshold, direction, and relative error criteria described above.  $TP_{\mathrm{up}}$ and $TP_{\mathrm{down}}$ denote the numbers of correctly predicted ramp-up and ramp-down events, respectively:
\begin{equation}
TP_{\mathrm{up}}
=
\left|
\left\{
i \;\middle|\;
r^{CSI}_i> \epsilon
\;\land\;
\hat{r}^{CSI}_i > \epsilon
\;\land\;
RE_i \leq \delta
\right\}
\right|
\end{equation}
\begin{equation}
TP_{\mathrm{down}}
=
\left|
\left\{
i \;\middle|\;
r^{CSI}_i < -\epsilon
\;\land\;
\hat{r}^{CSI}_i < -\epsilon
\;\land\;
RE_i \leq \delta
\right\}
\right|
\end{equation}

For this benchmark demonstration, we set $\delta=30\%$ as the evaluation threshold. 

\subsection*{Predictability and cloud regime analysis}

We first quantify site-level predictability using the rRMSE of the persistence model. This error reflects how strongly future irradiance or PV output departs from the current atmospheric state, with higher persistence rRMSE indicating lower predictability. To assess its consistency as an indicator of forecasting difficulty, we compute the Pearson correlation across sites between persistence rRMSE and the corresponding forecasting error of each evaluated model.

We then stratify the test data by cloud regimes to examine how atmospheric conditions contribute to differences in predictability across sites. Four regimes are considered --- clear, thin cloud, thick cloud, and patchy cloud. Clear conditions contain little or no visible cloud cover. Thin cloud conditions are characterized by high-level veil clouds or smooth, diffuse cloud layers that give the sky a milky appearance while the solar disk remains partly visible. Thick cloud conditions feature dense, opaque, and continuous overcast cloud cover, with no blue sky or solar disk visible. Patchy cloud conditions contain spatially heterogeneous or broken cloud fields with distinct cloudy and clear sky regions. This regime includes both stable translation of broken cloud structures and more dynamic conditions involving substantial structural evolution, multilayer motion, or the rapid entry of heterogeneous cloud fields through the image boundary. Additional examples illustrating the four cloud regimes across different SolarBench sites are provided in Supplementary Fig. 8.

For each site, we randomly select 1,000 test samples for manual cloud-regime labeling; when fewer than 1,000 test samples are available, all samples are labeled. Labels are assigned through visual inspection following a standardized sequence-level annotation guideline. Consistent with the model input configuration, each sample contains a sequence of four sky images captured at 5-min intervals over the 15-min input window. All four images are examined jointly to identify cloud regimes based on the spatial structure and temporal evolution of clouds across the sequence, including their opacity, spatial continuity, motion, and structural changes. Fisheye-lens distortion is considered when interpreting apparent changes in cloud size and shape to avoid confusing geometric distortion with physical cloud development. Labels are assigned using only the historical sky-image sequence, without reference to the corresponding target values, model predictions, or forecast errors. For each site and lead time, persistence rRMSE is then calculated separately within each cloud regime.

\subsection*{Data scaling and transfer learning}
We use the ViT model as the representative deep learning architecture for both the data scaling and transfer learning experiments. To quantify how much local training data are needed for ViT to outperform persistence under different site conditions, the model is trained at each site using progressively longer chronological subsets of the development data corresponding to 1, 3, 6, 12, and 24 months of equivalent local data, while the held-out test set is kept fixed. Data amounts are standardized by sample count, assuming 8 hours of daytime observations per day and a 5-min sampling interval. Because the datasets may contain missing observations and temporal gaps, the stated durations represent data-volume equivalents rather than contiguous calendar periods. The subsets are nested so that each larger subset extends the preceding subset with additional historical observations. All experiments use the same standardized task definitions and evaluation protocol. Forecast skill is calculated for each training data subset to assess how performance changes with local data availability.

To evaluate transfer learning across forecasting tasks and locations, we pre-train the ViT model for GHI forecasting and subsequently fine-tune it for PV power forecasting. We examine pre-training at three spatial scales using data from one, three, or six source sites. The source pool comprises six geographically distributed SolarBench sites with GHI observations: Golden, Folsom, Hong Kong, Sankt Augustin, Paris, and Nottingham. The pre-trained models are fine-tuned on two target PV systems: a 30 kW rooftop PV installation at Stanford University and a 3.275 MW hybrid PV system at the Gatton Solar Research Facility, University of Queensland. For single-site pre-training, separate models are trained using data from each of the six source sites. Based on their fine-tuning performance, the two best-performing source sites are retained. Each of the remaining four source sites is then combined with these two selected sites to construct four distinct three-site pre-training datasets. For six-site pre-training, data from all six source sites are combined into a single pre-training dataset.

Fine-tuning follows a warm-start strategy in which the ViT model is initialized with the pre-trained parameters and all parameters are subsequently updated using the available local PV training data. Local PV data availability is varied using progressively larger chronological subsets of the development data. For each local data subset, the fine-tuned models are compared with a ViT trained from scratch using the same amount of local PV data. A ViT trained from scratch using the full local development set is included as a full-data reference. All models are evaluated on the same held-out local test set using forecast skill.

\section*{Data availability}\label{sec7}
A sample of the SolarBench dataset is openly available at \url{https://huggingface.co/solarbench}. The full processed dataset will be made available to editors and referees upon request during peer review and will be released publicly upon publication through the same platform. 
The dataset is extensible, and future updates will be made publicly accessible through the same platform. To support proper attribution, SolarBench links each dataset component to its original source publication or DOI, where available. Users of data harmonized from third-party open datasets are encouraged to cite both SolarBench and the corresponding original dataset publications. For datasets released directly through SolarBench, citation of this work is sufficient. Dataset-specific citation instructions and source references are provided in the online documentation.

\section*{Code availability}\label{sec8}
The code for the SolarBench toolbox will be made available to editors and referees upon request during peer review and will be released publicly upon publication at \url{https://github.com/Solar4cast/solarbench} under the BSD-3-Clause license, with documentation provided at \url{https://solarbench.readthedocs.io}.

\section*{Acknowledgment}
This work was supported by Eni through its membership in the MIT Energy Initiative. The authors acknowledge the MIT Office of Research Computing and Data for providing high-performance computing resources that contributed to the research results reported in this paper. The authors thank the ESA $\Phi$-lab and Climate Office for their support during this research. The authors also acknowledge the following sources of funding for the collection and preparation of newly released individual ground-based datasets incorporated into SolarBench: the Engineering and Physical Sciences Research Council, UK (grant no. EP/W028581/1), the Department for Energy Security and Net Zero, UK (grant nos. 208857380 and 228516183), The Hong Kong Polytechnic University, and the German Federal Ministry of Education and Research (funding code 03SF0567A-G). The authors thank the project partners of the Father Franz Kruse Solar Energy Project, as well as the management and staff of St. Dominic’s Hospital in Akwatia and the Kologo Health Clinic, the citizens of Kologo, and the Ghanaian colleagues and partners who supported data collection at the two sites.

\bibliography{bibliography}

\end{document}











\section{Supplementary Notes}

This section provides detailed information on newly released ground-based datasets, as well as the atmospheric model products included in SolarBench, including climate reanalysis data (ERA5) and weather forecasts (GFS).

\subsection*{Supplementary Note 1. Newly Released Ground-based Datasets}

This section details raw data collection for the newly released ground-based datasets from Akwatia, Hong Kong, Kologo, Nottingham, and Sankt Augustin. Each dataset underwent initial preprocessing and quality control by the respective data contributors, as described for each site below. Before integration into the SolarBench database, all datasets underwent additional quality checks and standardized processing following the procedures described in the main paper.

\subsubsection*{Akwatia, Kologo, and Sankt Augustin}

The Akwatia and Kologo datasets were collected by the Energy Meteorology Lab of the International Centre for Sustainable Development at the University of Applied Sciences Bonn-Rhein-Sieg through the EnerSHelF project\footnote{\url{https://enershelf.de/}}, which supports the development of image-based solar nowcasting for PV applications in the Ghanaian health sector. Sankt Augustin serves as an additional German reference site and is located at the home institution of the contributing research group.

\paragraph{\textit{Akwatia and Kologo}}

\begin{figure}[H]
    \centering
    \includegraphics[width=1.\linewidth]{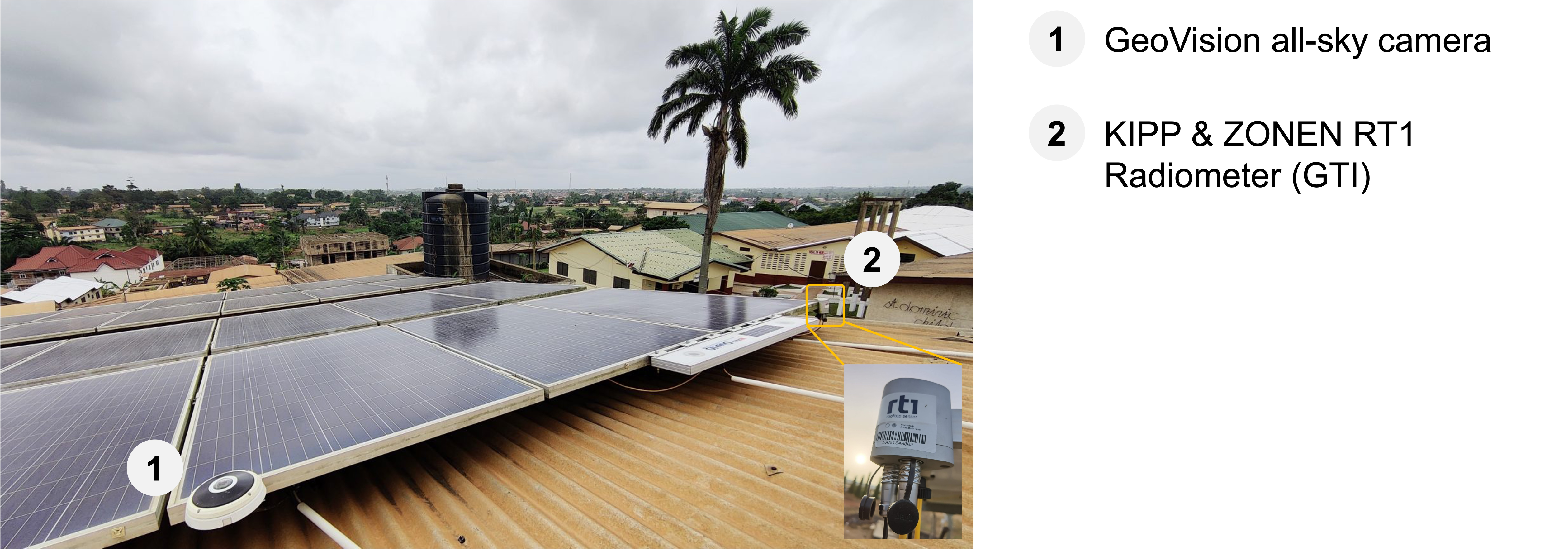}
    \caption{\textbf{Ground-based measurement device setup at the Akwatia site.} The measurement system consists of (1) a GeoVision all-sky camera and (2) a Kipp \& Zonen RT1 radiometer for GTI measurements. The inset shows a close-up view of the RT1 radiometer. The Kologo site uses the same measurement setup.}
    \label{fig:akwatia_devices}
\end{figure}

\noindent The Akwatia and Kologo datasets were collected at rural sites in southern Ghana (Akwatia: 6.0469$^\circ$ N, 0.7981$^\circ$ W; 147 m above sea level) and northern Ghana (Kologo: 10.7368$^\circ$ N, 1.0784$^\circ$ W; 163 m above sea level), respectively. Data were collected from April 2 to September 23, 2022, at Akwatia and from April 2 to October 4, 2022, at Kologo. The two sites used the same measurement setup (Supplementary Figure~\ref{fig:akwatia_devices}). At both sites, sky images were acquired at 1 Hz using a GeoVision GV-EFER3700-W all-sky camera with a 185$^\circ$ field of view and a native image resolution of 2048 $\times$ 1536 pixels. Global tilted irradiance (GTI) was measured at 1 Hz using a KIPP \& ZONEN RT1 Smart Rooftop Monitoring System, a radiometer operating over a spectral range of 400--1100 nm with a measurement resolution of 1 W m$^{-2}$. The sky imager and irradiance sensor were mounted with azimuth and zenith angles of 195$^\circ$ and 8$^\circ$, respectively, at Akwatia, and 181$^\circ$ and 8$^\circ$, respectively, at Kologo. As part of preprocessing and quality control, the sky images were cropped and resized to $448 \times 448$ pixels. For the irradiance measurements, days containing fewer than 100 missing samples were gap-filled using linear interpolation, whereas days with more than 100 missing samples were left uninterpolated and flagged. Further details of the preprocessing procedure are provided by Chaaraoui et al.~\cite{chaaraoui2025probabilistic}.

\paragraph{\textit{Sankt Augustin}}

\noindent The Sankt Augustin dataset was collected at a suburban site in Germany (50.7793$^\circ$ N, 7.1825$^\circ$ E; 65 m above sea level) from November 19, 2020 to November 14, 2022. The measurement instruments are shown in Supplementary Figure~\ref{fig:staugustin_devices}. Sky images were acquired at 1 Hz using a GeoVision GV-EFER3700-W all-sky camera with a 185$^\circ$ field of view and a native resolution of 2048 $\times$ 1536 pixels. Irradiance measurements were recorded every 10~s using a Kipp \& Zonen SOLYS2 sun tracker equipped with an SP Lite2 pyranometer for global horizontal irradiance (GHI), two CMP11 pyranometers for GHI and diffuse horizontal irradiance (DHI), respectively, and a GHP1 pyrheliometer for direct normal irradiance (DNI). As part of preprocessing and quality control, the sky images were cropped and resized to $448 \times 448$ pixels.

Within SolarBench, the GHI measurements from the SP Lite2 pyranometer are used as the standard GHI variable because its silicon sensor has a spectral response that more closely matches those of photovoltaic panels and the all-sky camera than the thermopile-based CMP11 pyranometer. The CMP11 GHI measurements are retained as an additional variable.

\subsubsection*{Hong Kong}

The Hong Kong dataset was collected by the Renewable Energy Advancement Lab at The Hong Kong Polytechnic University ($22.3047^\circ$N, $114.1803^\circ$E) from February 18 to December 31, 2023. The dataset includes all-sky images, solar irradiance measurements, and surface meteorological observations at a temporal resolution of 1 min. The measurement instruments are shown in Supplementary Figure~\ref{fig:hk_polyu_devices}. All-sky images were acquired continuously using a Vivotek FE9380-HV network camera. The raw images have a spatial resolution of $512 \times 512$ pixels and were recorded throughout the full 24-hour daily cycle. Solar irradiance and meteorological observations were recorded using a Delta-T SPN1 pyranometer and a Lufft WS500 weather station co-located with the camera. The SPN1 measured GHI and DHI, from which DNI was calculated. The weather station recorded atmospheric pressure, air temperature, relative humidity, wind speed, and wind direction. All variables were provided as 1-min averages. To preserve the original high-resolution observations, no additional quality control procedures were applied before the dataset was provided to SolarBench. 

\subsubsection*{Nottingham}

The Nottingham dataset was collected on the campus of the University of Nottingham ($52.9521 ^\circ$N, $1.1843 ^\circ$W) from January 5 to December 31, 2023. The dataset includes all-sky images, solar irradiance measurements, and surface meteorological observations at a temporal resolution of 1 min. The measurement instruments are shown in Supplementary Figure~\ref{fig:unottingham_devices}. All-sky images were acquired using a Mobotix Q26B fisheye camera with a native image resolution of $1280 \times 1280$ pixels. During preprocessing, approximately 30 pixels were removed from each side to center the circular field of view before the images were resized to $128 \times 128$ pixels. Solar irradiance measurements were obtained using a Kipp \& Zonen RaZON+ sun tracker, which measures DNI and DHI, from which GHI is derived using the corresponding solar zenith angle. An independent Kipp \& Zonen CMP10 pyranometer simultaneously measured GHI for cross-validation. Surface meteorological observations, including air temperature, relative humidity, atmospheric pressure, wind speed, and wind direction were recorded using a Gill Instruments GMX600 weather station. 

The irradiance measurements underwent several quality control filters following the procedures described by Yang et al.~\cite{yang2020ensemble}. First, extremely rare irradiance values were removed following Long and Shi~\cite{long2008automated}:
\begin{align}
    \mathrm{GHI} &< 1.2 E_{0n}\cos^{1.2}(Z) + 50, \\
    \mathrm{DHI} &< 0.75 E_{0n}\cos^{1.2}(Z) + 30, \\
    \mathrm{DNI} &< 0.95 E_{0n}\cos^{0.2}(Z) + 10,
\end{align}
where $E_{0n}$ is the extraterrestrial irradiance on a surface normal to the solar beam and $Z$ is the solar zenith angle.

Second, diffuse ratio tests were then applied following Long and Shi~\cite{long2008automated}:
\begin{align}
\frac{\mathrm{DHI}}{\mathrm{GHI}} &< 1.05,
\text{for}\ Z < 75^\circ \ \text{and}\ \mathrm{GHI} > 50, \\
\frac{\mathrm{DHI}}{\mathrm{GHI}} &< 1.10,  \text{for}\ Z > 75^\circ \ \text{and}\ \mathrm{GHI} > 50.
\end{align}

Additional filters followed Gueymard and Ruiz-Arias~\cite{gueymard2016extensive}:
\begin{align}
Z &< 75^\circ, \\
\mathrm{GHI} &> 0\ \text{and}\ \mathrm{DHI} > 0, \\
-1 &< \mathrm{DNI} < 1100 + 0.03h, \\
\left|\mathrm{closr}\right| & < 0.05.
\end{align}
where $h$ is the station elevation in meters. The irradiance component closure ratio was defined as
\begin{equation}
\mathrm{closr}
=
\frac{\mathrm{GHI}-\mathrm{DHI}-\mathrm{DNI}\cos(Z)}
{\mathrm{GHI}},
\end{equation}

The original solar zenith angle threshold of $85^\circ$ was reduced to $75^\circ$ to account for the effective coverage of the fisheye camera. 

In the above quality control steps, all irradiances are expressed in $\mathrm{W\ m^{-2}}$. The instruments and optical surfaces were inspected and cleaned regularly throughout the measurement campaign. A tower located southeast of the measurement station casts shadows at low solar elevations, particularly during afternoons from November to March. These periods were retained in the released dataset and no additional filtering was applied.

\subsection*{Supplementary Note 2. Atmospheric Model Products}

SolarBench includes two atmospheric model products as complementary inputs for solar nowcasting: the ECMWF Reanalysis v5 (ERA5) climate reanalysis~\cite{hersbach2020} and weather forecasts from the Global Forecast System (GFS)~\cite{CLOUGH2005233}.

\subsubsection*{ERA5}

ERA5 is a global atmospheric reanalysis produced by the European Centre for Medium-Range Weather Forecasts (ECMWF). It combines numerical weather models with assimilated observations to provide spatially and temporally consistent estimates of atmospheric conditions. The standard ERA5 product is available at an hourly temporal resolution on a $0.25^{\circ}\times0.25^{\circ}$ spatial grid, corresponding to $31~\mbox{km}/\text{pixel}$ at the equator, with records extending from 1950 to present. 

Supplementary Table~\ref{tab:era5_variables} lists the ERA5 variables processed and included in the SolarBench database. For each site, the ERA5 data are provided for the period corresponding to the available ground-based observations. Most variables were obtained from the European Union's Earth Data Hub platform~\footnote{Earth Data Hub -- \url{https://earthdatahub.destine.eu/}}, whereas \textit{10m wind gust since previous post-processing}, \textit{instantaneous 10m wind gust} and \textit{cloud base height} were obtained from the ECMWF Climate Data Store~\footnote{ECMWF Climate Data Store -- \url{https://cds.climate.copernicus.eu/datasets/reanalysis-era5-single-levels?tab=download}}. 

The ERA5 data are provided for each site in two different forms: the single nearest grid point to each site and the four nearest grid points forming a 2$\times$2 grid around the site.

\subsubsection*{GFS}

GFS is a global numerical weather prediction system operated by the National Centers for Environmental Prediction (NCEP). It uses a coupled atmosphere–ocean–land modeling system and assimilates meteorological observations to generate forecasts of atmospheric and surface conditions. The highest-resolution global product is provided on an approximately \(0.25^{\circ} \times 0.25^{\circ}\) grid, corresponding to about \(31~\mbox{km}/\text{pixel}\) at the equator. Forecasts are initialized every 6 h, with the available forecast horizon and temporal interval varying over the data record.

The forecast configuration varies by date:
\begin{itemize}
    \item From 2015/04/01-2017/07/09: 36-hour forecasts, excluding hour 0, at 3-hour intervals.
    \item From 2017/07/09-2021/06/11: 384-hour forecasts, at 1-hour intervals from hours 0-120, at 3-hour intervals from hours 120-240, and 12-hour intervals from hours 240-384.
    \item From 2021/06/12: 384-hour forecasts, at 1-hour intervals from hours 0-120 and 3-hour intervals from hours 120-384.
\end{itemize}

Supplementary Table~\ref{tab:gfs_variables} lists the GFS variables included in SolarBench. For each location, GFS data are provided over the period corresponding to the available ground-based observations, when available. Forecasts initialized every 6 h are retained up to a lead time of 36 h. The data were obtained through Google Earth Engine (GEE)~\footnote{GEE -- \url{https://developers.google.com/earth-engine/datasets/catalog/NOAA_GFS0P25}}.

The GFS data are provided for each location in four forms: (1) the single nearest grid point, (2) the four nearest grid points forming a 2$\times$2 grid,  (3) bilinearly interpolated values from the 16 nearest grid points, and (4) bicubically interpolated values from the 16 nearest grid points.

\section*{Supplementary Methods}

This section provides additional implementation details for the deep learning models evaluated in the SolarBench benchmarking experiments, including their architectures, input configurations, and training hyperparameters. It also describes evaluation metrics implemented in the SolarBench toolbox beyond those reported in the main paper.

\subsection*{Model Implementation and Training Details}

All deep learning models were implemented using PyTorch v2.3.1 within the SolarBench toolbox. Unless otherwise stated, models use observations from the preceding 15 min sampled at 5-min intervals and are trained separately for each forecast lead time. The main architectural characteristics of each model are described in the Methods; here, we document implementation sources, model-specific adaptations, and training configurations used in the benchmark experiments.

\subsubsection*{Multilayer Perceptron}

The multilayer perceptron (MLP) follows the architecture of Pedro et al.~\cite{pedro2012assessment} and uses only historical irradiance or PV output measurements as input. The model was trained using the Adam optimizer with a learning rate of $3\times10^{-4}$, a batch size of 256, and the mean squared error loss. Training was initially scheduled for 40 epochs. If early stopping was not triggered within this period, training was continued until the validation loss failed to improve for five consecutive epochs. The checkpoint with the lowest validation loss was retained.

\subsubsection*{SUNSET}

SUNSET was originally developed by Sun et al.~\cite{sun2019short}. We adopted the publicly available TensorFlow implementation \footnote{\url{https://github.com/yuhao-nie/Stanford-solar-forecasting-dataset}} and reimplemented it in PyTorch for integration into the SolarBench toolbox. Compared with the original implementation, the input sequence is sampled at 5-min intervals over the preceding 15 min rather than at 1-min intervals.

The model was trained using the Adam optimizer, a learning rate of $3 \times 10^{-6}$, a batch size of 256, and the mean squared error loss. Training was initially scheduled for 40 epochs. If early stopping was not triggered within this period, training was continued until the validation loss failed to improve for five consecutive epochs. The checkpoint with the lowest validation loss was retained.

\subsubsection*{OmniVision}

OmniVision was introduced by Paletta et al.~\cite{paletta2023omnivision}. Our implementation follows the architecture described in the paper and adapts the publicly available PyTorch implementation \footnote{\url{https://github.com/tcapelle/eclipse_pytorch}} of the Eclipse model \cite{paletta2022eclipse}, which uses a similar spatiotemporal backbone but does not include a satellite image branch. 

Compared with the original OmniVision and Eclipse formulations, we remove the iterative future state prediction procedure and disable the future image segmentation objective. Instead, a separate model directly predicts irradiance or PV output at each evaluated lead time.

The model was trained using the AdamW optimizer with an initial learning rate of $1 \times 10^{-4}$, a weight decay of $1 \times 10^{-4}$, and a batch size of 128, and the mean squared error loss. The learning rate was linearly increased during a three-epoch warm-up phase and subsequently reduced using cosine annealing. Training was initially scheduled for 40 epochs. If early stopping was not triggered within this period, training was continued until the validation loss failed to improve for ten consecutive epochs. The checkpoint with the lowest validation loss was retained.

\subsubsection*{Vision Transformer with Temporal Attention}

The Vision Transformer (ViT) model is based on the implementation contributed by the authors of Zhang et al.~\cite{zhang2023advanced}. The original model uses the most recent sky image together with irradiance and other meteorological measurements. We instead use a sequence of historical sky images and retain only the corresponding irradiance or PV output measurements as numerical inputs. Temporal encoding inspired by TimeSFormer~\cite{bertasius2021space} is introduced to jointly model the image and measurement sequences.

The model was trained using the Adam optimizer with a learning rate of $3\times10^{-4}$, a weight decay of $1\times10^{-4}$, a batch size of 32, and the mean squared error loss. Training was initially scheduled for 20 epochs using a cosine learning-rate schedule that reduced the learning rate to 10\% of its initial value. If early stopping was not triggered within this period, training was continued until the validation loss failed to improve for 20 consecutive epochs. The checkpoint with the lowest validation loss was retained.

\subsection*{Additional Evaluation Metrics}

In addition to RMSE, rRMSE, forecast skill, and ramp recall described in the main paper, the SolarBench toolbox implements additional metrics for evaluating forecast magnitude, systematic bias, and temporal misalignment. Let $y_i$ and $\hat{y}_i$ denote the observed and predicted values, respectively, for $i=1,\ldots,N$.

\subsubsection*{Mean Absolute Error}

The mean absolute error (MAE) measures the average magnitude of the forecast errors without considering their direction:
\begin{equation}
\mathrm{MAE}
=
\frac{1}{N}
\sum_{i=1}^{N}
\left|
\hat{y}_i-y_i
\right|.
\end{equation}
Compared with RMSE, MAE is less sensitive to a small number of large forecast errors. A lower MAE indicates better forecast accuracy.

\subsubsection*{Mean Bias Error}

The mean bias error (MBE) measures systematic overprediction or underprediction:
\begin{equation}
\mathrm{MBE}
=
\frac{1}{N}
\sum_{i=1}^{N}
\left(
\hat{y}_i-y_i
\right).
\end{equation}
A positive MBE indicates that the model overpredicts the target variable on average, whereas a negative MBE indicates systematic underprediction. An MBE close to zero indicates little overall bias, although positive and negative errors may cancel.

\subsubsection*{Temporal Distortion Mix}

The SolarBench toolbox also includes an experimental implementation of the temporal distortion mix (TDM) \cite{vallance2017towards}, which assesses the direction of temporal misalignment between predicted and observed time series. TDM is not evaluated in this study, and its implementation has not yet been independently validated. Further implementation details and usage guidance are provided in the SolarBench online documentation \url{https://solarbench.readthedocs.io}.

\section*{Supplementary Figures}

\begin{figure}[H]
    \centering
    \includegraphics[width=1.\linewidth]{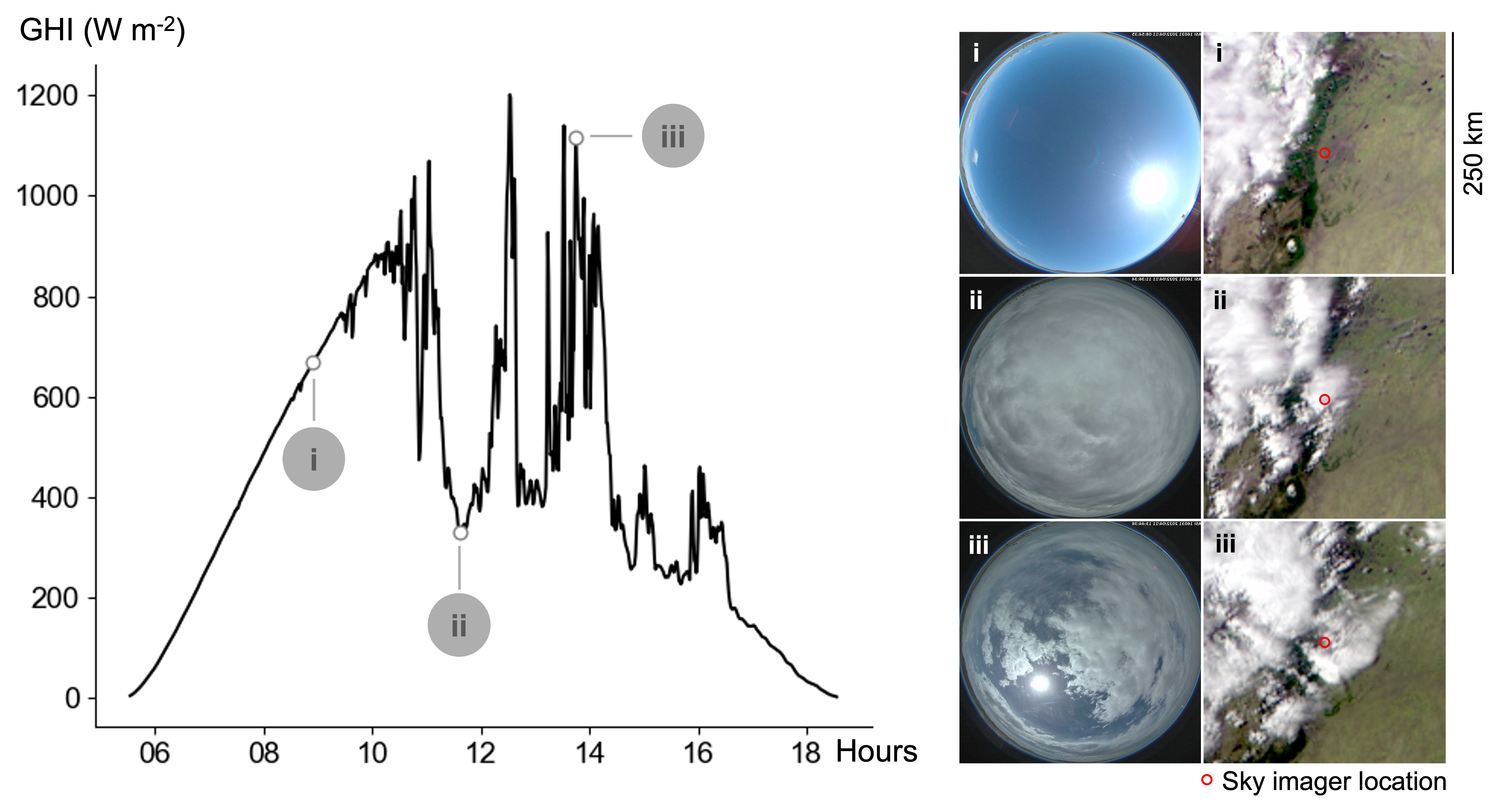}
    \caption{\textbf{Example solar irradiance variability and corresponding cloud cover observations}. Left: measured GHI over a representative day. Three time points (i–iii) representing distinct irradiance conditions are highlighted. Right: corresponding ground-based sky images (left column) and geostationary satellite images (right column). The red circle indicates the location of the sky imaging system within each satellite scene. These examples illustrate how cloud observations from ground-based sky cameras and geostationary satellites provide information about short-term variability in surface solar irradiance.}
    \label{fig:sky_sat_imagery_sample}
\end{figure}

\begin{figure}[H]
    \centering
    \includegraphics[width=.99\linewidth]{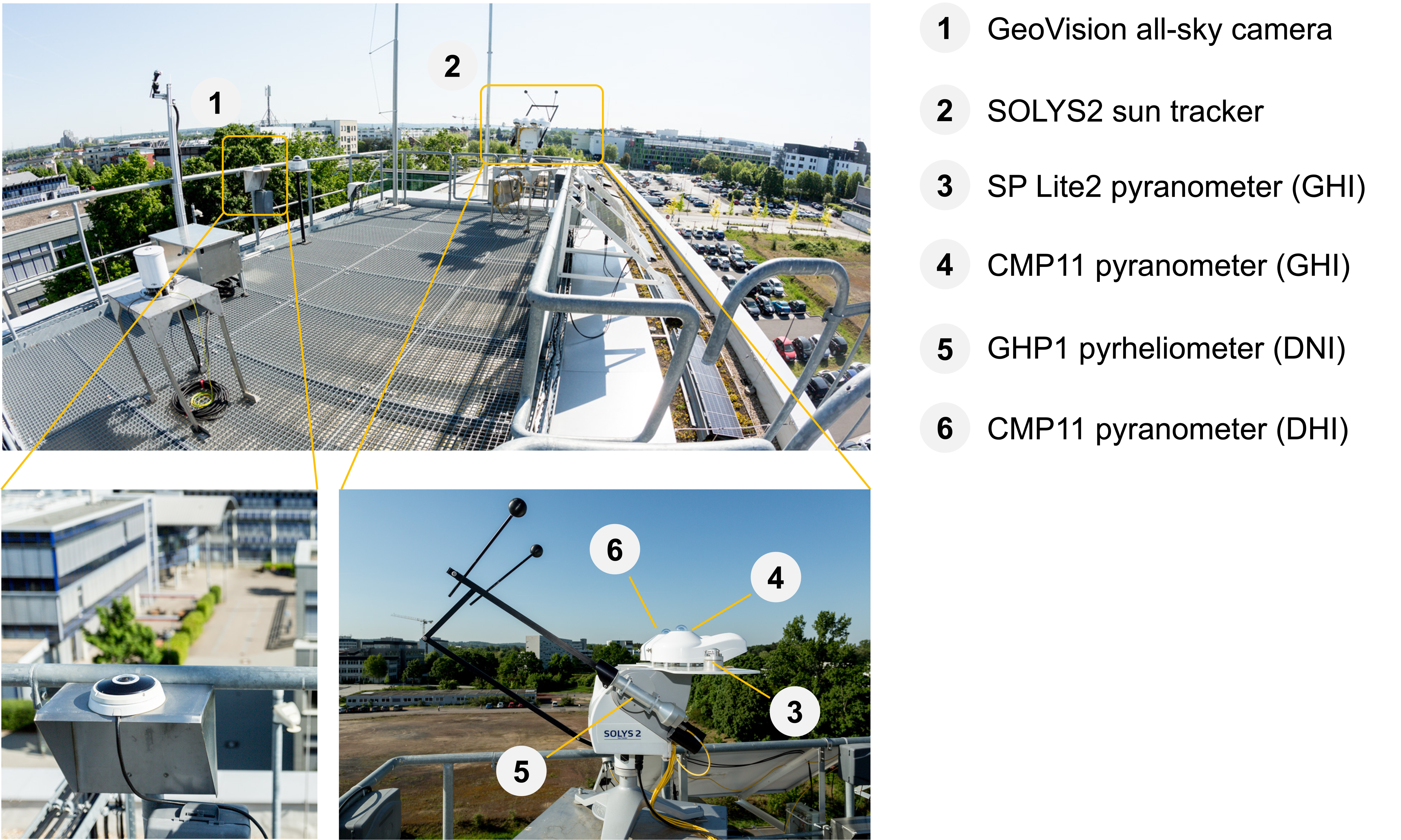}
    \caption{\textbf{Ground-based measurement device setup at the Sankt Augustin site.} The system includes (1) a GeoVision all-sky camera, (2) a SOLYS2 sun tracker, (3) an SP Lite2 pyranometer for global horizontal irradiance (GHI) measurements, (4) a CMP11 pyranometer for global horizontal irradiance (GHI) measurements, (5) a GHP1 pyrheliometer for direct normal irradiance (DNI) measurements, and (6) a CMP11 pyranometer for diffuse horizontal irradiance (DHI) measurements. Instruments (3)–(6) are mounted on the SOLYS2 sun tracker. Enlarged views of the all-sky camera and the solar radiation measurement system are shown in the lower panels.}
    \label{fig:staugustin_devices}
\end{figure}

\begin{figure}[H]
    \centering
    \includegraphics[width=1.\linewidth]{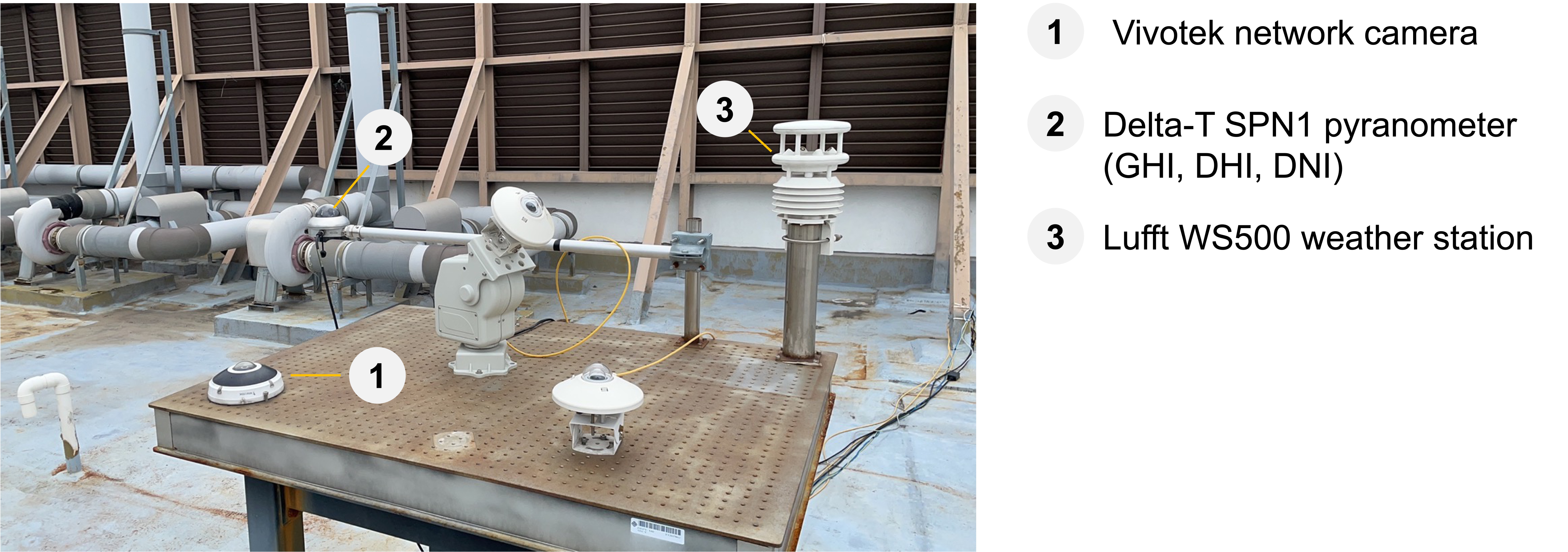}
    \caption{Ground-based measurement device setup at the Hong Kong site. The system includes (1) a Vivotek FE9380-HV network camera for capturing sky images, (2) a Delta-T SPN1 pyranometer measuring GHI and DHI, from which DNI is calculated, and (3) a Lufft WS500 weather station providing measurements of air temperature, relative humidity, atmospheric pressure, wind speed, and wind direction.} 
    \label{fig:hk_polyu_devices}
\end{figure}

\begin{figure}[H]
    \centering
    \includegraphics[width=1.\linewidth]{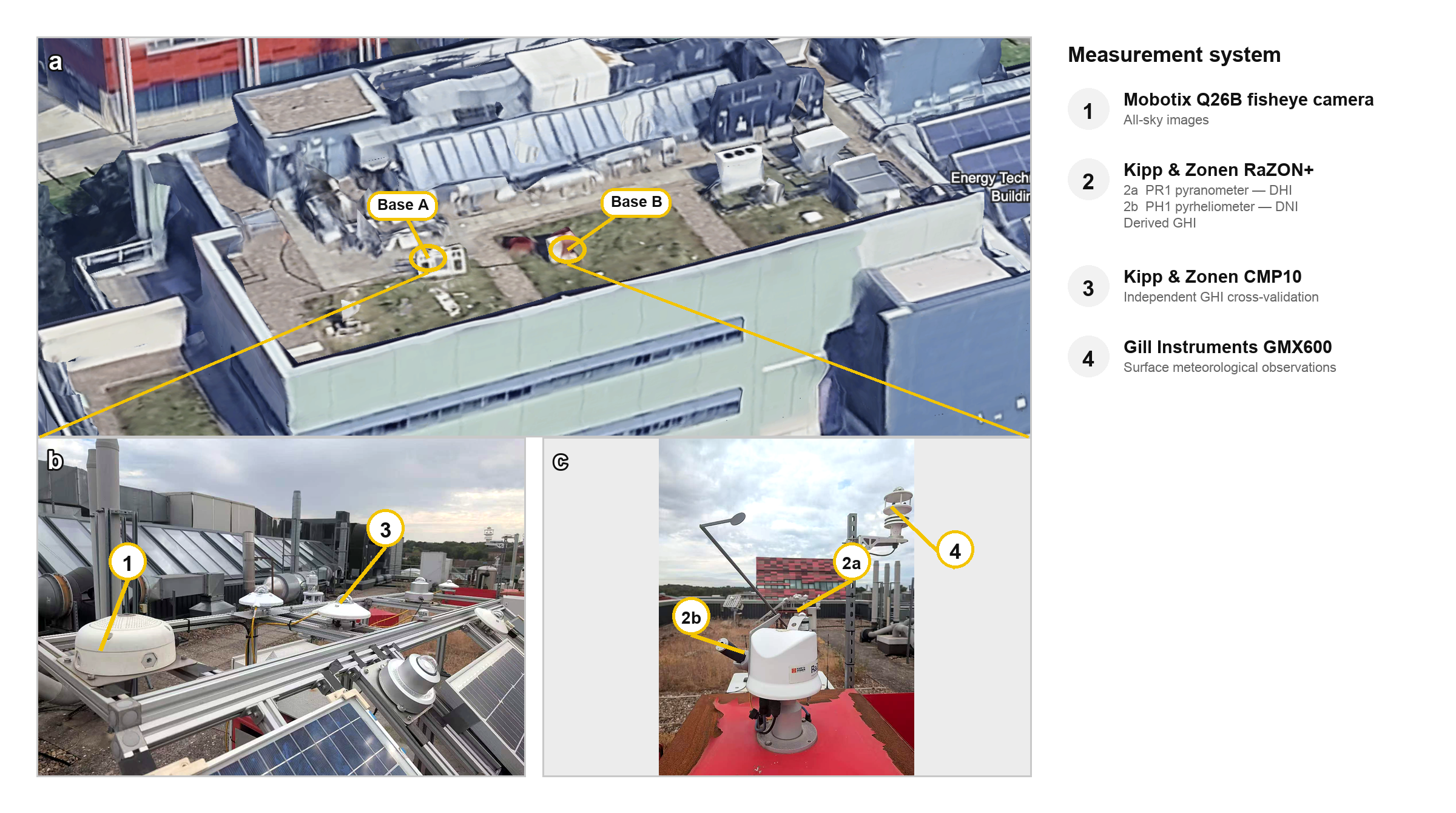}
    \caption{\textbf{Ground-based measurement device setup at the Nottingham site.} Panel (a) shows the rooftop site at the University of Nottingham's Jubilee Campus and the locations of Base A and Base B, enlarged in panels (b) and (c). The system includes (1) a Mobotix Q26B fisheye camera, (2) a Kipp \& Zonen RaZON+ sun tracker with a PR1 pyranometer for DHI and a PH1 pyrheliometer for DNI, from which GHI is derived, (3) a CMP10 pyranometer for GHI cross-validation, and (4) a Gill Instruments GMX600 weather station.}
    \label{fig:unottingham_devices}
\end{figure}

\begin{figure}[H]
    \centering
    \includegraphics[width=1.\linewidth]{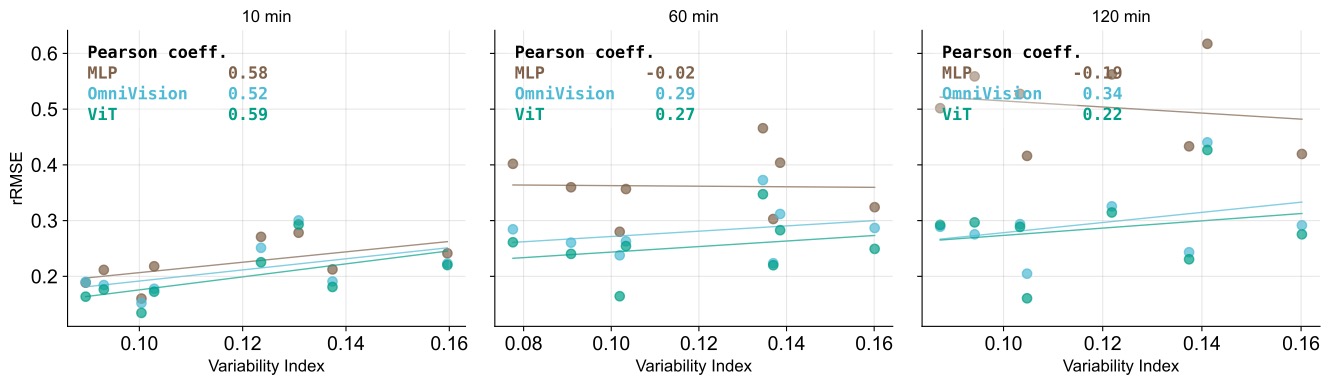}
    \caption{\textbf{Relationship between forecasting error and the variability index.} rRMSE of the MLP, OmniVision, and ViT models is plotted against the variability index for the 10-, 60-, and 120-min lead times. Each point represents one SolarBench site, and solid lines indicate least-squares linear fits. Pearson correlation coefficients between rRMSE and the variability index are reported for each model. The variability index is computed using a fixed 5-min interval following Stein et al. \cite{stein2012variability} for all forecast lead times. The weak and inconsistent correlations across models and lead times indicate that this aggregate variability metric alone does not adequately characterize forecasting difficulty.}
    \label{fig:forecast_error_vs_vi}
\end{figure}

\begin{figure}[H]
    \centering
    \includegraphics[width=1.\linewidth]{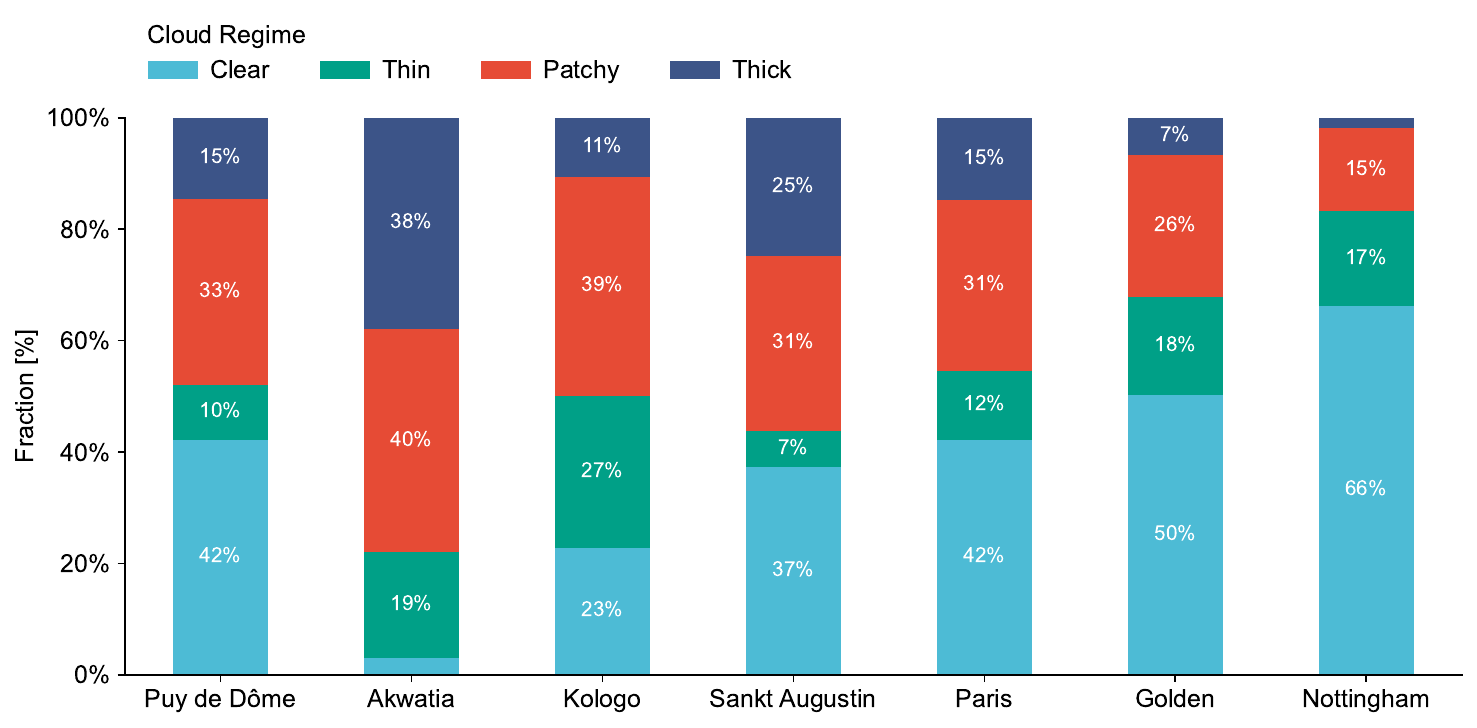}
    \caption{\textbf{Distribution of cloud regimes across SolarBench sites.} Stacked bar charts showing the fraction of samples assigned to the four cloud regimes (Clear, Thin, Patchy, and Thick) at each site. Percentages denote the proportion of samples in each regime, highlighting substantial differences in cloud regime composition across the benchmark locations.}
    \label{fig:cloud_regime_dist}
\end{figure}

\begin{figure}[H]
    \centering
    \includegraphics[width=1.\linewidth]{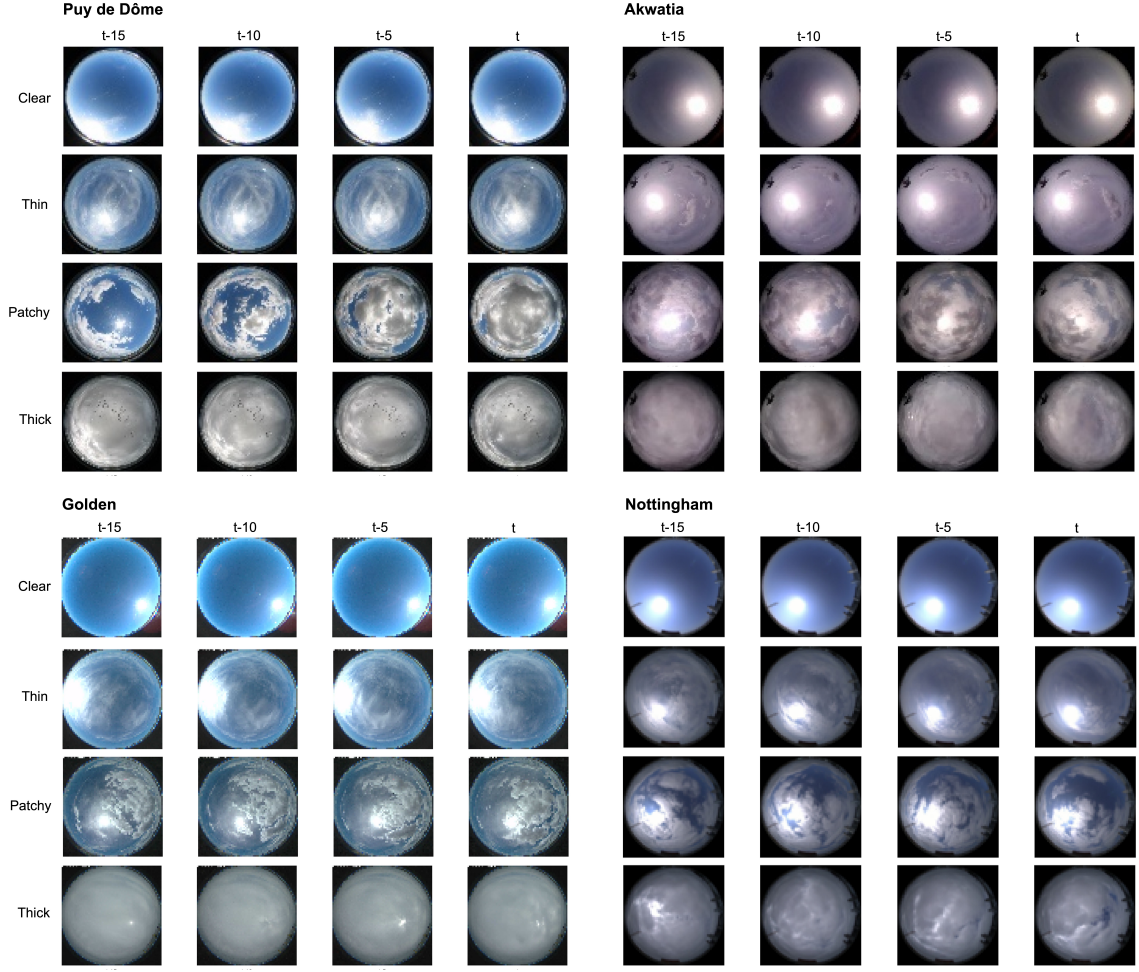}
    \caption{\textbf{Representative cloud regime examples across SolarBench sites.} Sky image sequences illustrating clear, thin cloud, patchy cloud, and thick cloud conditions at Puy de Dôme, Akwatia, Golden, and Nottingham. Each example shows a 15-min historical sequence at 5-min intervals ($t-15$, $t-10$, $t-5$, and $t$), highlighting cross-site differences in the visual appearance of each cloud regime.}
    \label{fig:cloud_regime_examples}
\end{figure}

\begin{landscape}
\section*{Supplementary Tables}

\begin{table}[H]
\centering
\caption{\textbf{Site Characteristics.}}
\label{tab:site_charac}
\scriptsize
    \begin{tabularx}{\linewidth}{l | p{2.6cm} | l | l | l | X }
\toprule
\textbf{Site} & \textbf{Dataset Name (on Hugging Face)} & \textbf{Lat., Lon. ($^\circ$)}& \textbf{Köppen--Geiger Climate Class} & \textbf{Data Period} & \textbf{Duration (years)} \\
\midrule
Akwatia, Ghana & \texttt{EnerSHelF-Akwatia} & 6.047, -0.797 & Aw: Tropical savanna with dry winters & 2022-04-02 -- 2022-09-23 & 0.5 \\
Folsom, USA & \texttt{UCSD-Folsom} & 38.642, -121.148 & Csa: Hot-summer Mediterranean & 2014-01-02 -- 2016-12-30 & 3.0 \\
Gatton, Australia & \texttt{UQueensland}  & -27.559, 152.343 & Cfa: Humid subtropical & 2020-01-23 -- 2020-05-06 & 0.3 \\
Golden, USA & \texttt{SRRL-BMS} & 39.742, -105.180 & Dfb: Warm-summer humid continental & 2021-04-17 -- 2024-03-22 & 2.9 \\
Hong Kong, China  & \texttt{HK-PolyU} & 22.305, 114.180 & Cwa: Dry-winter humid subtropical & 2023-02-18 -- 2023-12-31  & 0.9 \\
Kologo, Ghana & \texttt{EnerSHelF-Kologo} &  10.737, -1.078 & Aw: Tropical savanna with dry winters & 2022-04-02 -- 2022-10-04 & 0.5  \\
Nottingham, UK & \texttt{UNottingham} & 52.952, -1.184 & Cfb: Temperate oceanic  & 2023-01-05 -- 2023-12-31 & 1.0\\
Paris, France & \texttt{SIRTA} & 48.713, 2.208 & Cfb: Temperate oceanic  & 2015-01-02 -- 2022-12-31 & 8.0 \\
Puy de Dôme, France & \texttt{COPDD} & 45.774, 2.962 & Cfb: Temperate oceanic  & 2015-12-01 -- 2023-08-27 & 7.7\\
Sankt Augustin, Germany & \texttt{EnerSHelF-StAugustin} & 50.779, 7.183 & Cfb: Temperate oceanic  & 2020-11-19 -- 2022-11-14 & 2.0\\
Stanford, USA & \texttt{SKIPPD} & 37.427, -122.174 & Csb: Warm-summer Mediterranean & 2017-03-09 -- 2019-10-26  & 2.6 \\
\bottomrule
\end{tabularx}
\end{table}

\begin{table}[H]
\centering
\begin{threeparttable}
\caption{Details of the ground-based dataset for each site.}
\label{tab:ground_data_details}
\scriptsize
\begin{tabularx}{\linewidth}{l | l |  p{1.cm} | p{.7cm} | l | X  }
\toprule
\textbf{Site} & \textbf{Sources} & 
\textbf{Temp. Res. (min)}& \textbf{Size (GB)}& 
\textbf{Samples} & \textbf{Available Data} \\
\midrule
Akwatia, Ghana & SolarBench & 1 & 0.6 & 89,953 & Sky image, GTI \\
Folsom, USA & \cite{pedro2019comprehensive} & 1 & 5.9 & 765,140 & Sky image, GHI, DNI, DHI, Air temperature, Relative humidity, Pressure, Wind speed, Wind direction, Max wind speed, Precipitation  \\
Gatton, Australia & \cite{xu2023gatton} & 1 & 0.4 & 69,347 &  Sky image, PV output \\
Golden, USA & \cite{andreas1981srrl} & 1 & 2.9 & 367,878 & Sky image, GHI, DNI, DHI, Air temperature, Relative humidity, Pressure, Wind speed, Wind direction, Max wind speed, Precipitation \\
Hong Kong, China  & SolarBench & 1 & 1.5 & 223,812 & Sky image, GHI, DHI, Air temperature, Relative humidity, Pressure, Wind speed, Wind direction  \\
Kologo, Ghana & SolarBench & 1 & 0.7 & 92,182 & Sky image, GTI  \\
Nottingham, UK & SolarBench & 1 & 0.5 & 68,237 &  Sky image, GHI, DNI, DHI, GHI\_min\_avg, DNI\_min\_avg, DHI\_min\_avg, Clear-sky GHI, Clear-sky DNI, Clear-sky DHI, Air temperature, Relative humidity, Pressure, Wind speed, Wind direction, Solar zenith angle, Solar azimuth angle \\
Paris, France & \cite{haeffelin2005sirta} & 1$\sim$2$^1$ & 10.5 & 1,368,060 &  Sky image, GHI, GHI\_std, GHI\_min, GHI\_max, DNI, DNI\_std, DNI\_min, DNI\_max, DHI, DHI\_std, DHI\_min, DHI\_max, LWd\_max, Air temperature, Relative humidity, Pressure \\
Puy de Dôme, France  & \cite{baray2020cezeaux} & 1$\sim$2$^2$ & 7.4 & 981,409 & Sky image, GHI, DNI, DHI, Air temperature, Relative humidity, Pressure, Wind speed, Wind direction, Precipitation  \\
Sankt Augustin, Germany & SolarBench & 1 & 3.3 & 442,788 & Sky image, GHI, DNI, DHI  \\
Stanford, USA  & \cite{nie2023skipp} & 1 & 2.3 & 363,375 & Sky image, PV output  \\
\bottomrule
\end{tabularx}
\begin{tablenotes}[flushleft]
\scriptsize
\item[1] 2015-01 -- 2017-12: 1 min; 2018-01 -- 2022-12: 2 min
\item[2] 2015-12 -- 2018-08: 1 min; 2018-10 -- 2023-08: 2 min
\end{tablenotes}

\end{threeparttable}
\end{table}

\begin{table}[H]
\centering
\begin{threeparttable}
\caption{Details of geostationary satellite imagery data for each site.}
\label{tab:geo_sat_details}
\scriptsize
\begin{tabularx}{\linewidth}{l | l | p{1.cm} | p{.7cm} | l| l | X }
\toprule
\textbf{Site} & \textbf{Sources} & 
\textbf{Temp. Res. (min)} &
\textbf{Size (GB)}& \textbf{Samples} & \textbf{Bands} & \textbf{Access link} \\
\midrule
Akwatia, Ghana & MSG & 15 & 3.3 & 13,720 & 12 & \url{https://user.eumetsat.int/resources/user-guides/eumetsat-data-access-client-eumdac-guide} \& \url{https://github.com/openclimatefix/satellite-consumer} \\
Folsom, USA & N/A & N/A & N/A & N/A & N/A & N/A \\
Gatton, Australia & Himawari-8 & 10 & 3.3 & 9,833 & 16 & \url{https://noaa-himawari8.s3.amazonaws.com/index.html\#AHI-L1b-FLDK/}$^1$ \\
Golden, USA & GOES-16 & 5 & 48.5 & 118,598 & 16 & \url{https://noaa-goes16.s3.amazonaws.com/index.html\#ABI-L1b-RadC/} \\
Hong Kong, China  & Himawari-9 & 10 & 9.5 & 29,305 & 16 & \url{https://noaa-himawari9.s3.amazonaws.com/index.html\#AHI-L1b-FLDK/} $^2$\\
Kologo, Ghana & MSG & 15 & 3.4 & 13,882 & 12 & \url{https://user.eumetsat.int/resources/user-guides/eumetsat-data-access-client-eumdac-guide} \& \url{https://github.com/openclimatefix/satellite-consumer} \\
Nottingham, UK & MSG & 5 & 20.2 & 89,208 & 12 & \url{https://user.eumetsat.int/resources/user-guides/eumetsat-data-access-client-eumdac-guide} \& \url{https://github.com/openclimatefix/satellite-consumer} \\
Paris, France & MSG & 5 & 163 & 742,522 & 12 & \url{https://user.eumetsat.int/resources/user-guides/eumetsat-data-access-client-eumdac-guide} \& \url{https://github.com/openclimatefix/satellite-consumer} \\
Puy de Dôme, France  & MSG & 5 & 163 & 716,267 & 12 & \url{https://user.eumetsat.int/resources/user-guides/eumetsat-data-access-client-eumdac-guide} \& \url{https://github.com/openclimatefix/satellite-consumer} \\
Sankt Augustin, Germany & MSG & 5 & 39.3 & 179,143 & 12 & \url{https://user.eumetsat.int/resources/user-guides/eumetsat-data-access-client-eumdac-guide} \& \url{https://github.com/openclimatefix/satellite-consumer} \\
Stanford, USA  & N/A & N/A & N/A & N/A & N/A & N/A\\
\bottomrule
\end{tabularx}
\begin{tablenotes}[flushleft]
\scriptsize
\item[1] Segment: S0810
\item[2] Segment: S0301
\end{tablenotes}

\end{threeparttable}
\end{table}
\end{landscape}

\begin{table}[H]
    \scriptsize
    \caption{Processed ERA5 variables.}
    \label{tab:era5_variables}
    \centering
    \begin{tabularx}{\textwidth}{l|l|l}
    \toprule
        \textbf{Description} & \textbf{Variable Name} & \textbf{Unit} \\ \midrule
Total precipitation & \texttt{tp} & mm\\
10m U wind component & \texttt{u10} & m/s\\
10m V wind component & \texttt{v10} & m/s\\
100m V wind component & \texttt{v100} & m/s\\
100m U wind component & \texttt{u100} & m/s\\
2m temperature & \texttt{t2m} & K\\
2m dew point temperature & \texttt{d2m} & K\\
Top net short-wave (solar) radiation, clear sky & \texttt{tsrc} & J m$^{\text{-2}}$\\
Top net short-wave (solar) radiation & \texttt{tsr} & J m$^{\text{-2}}$\\
Surface short-wave (solar) radiation downwards & \texttt{ssrd} & J m$^{\text{-2}}$\\
Surface net short-wave (solar) radiation & \texttt{ssr} & J m$^{\text{-2}}$\\
Total column cloud ice water & \texttt{tciw} & kg m$^{\text{-2}}$\\
Total column cloud liquid water & \texttt{tclw} & kg m$^{\text{-2}}$\\
High cloud cover & \texttt{hcc} & (0 - 1)\\
Medium cloud cover & \texttt{mcc} & (0 - 1)\\
Low cloud cover & \texttt{lcc} & (0 - 1)\\
Total cloud cover & \texttt{tcc} & (0 - 1)\\
Surface direct short-wave radiation, clear sky & \texttt{cdir} & J m$^{\text{-2}}$\\
Total sky direct short-wave (solar) radiation at surface & \texttt{fdir} & J m$^{\text{-2}}$\\
Surface net short-wave (solar) radiation, clear sky & \texttt{ssrc} & J m$^{\text{-2}}$\\
Surface short-wave (solar) radiation downward clear-sky & \texttt{ssrdc} & J m$^{\text{-2}}$\\
Top of atmosphere incident short-wave (solar) radiation & \texttt{tisr} & J m$^{\text{-2}}$\\
Forecast albedo & \texttt{fal} & (0 - 1)\\
10m wind gust since previous post-processing & \texttt{10m\_wind\_gust\_since\_previous\_post\_processing} & m/s\\
Instantaneous 10m wind gust & \texttt{instantaneous\_10m\_wind\_gust} & m/s\\
Cloud base height & \texttt{cloud\_base\_height} & m\\
\bottomrule
    \end{tabularx}
\end{table}

\begin{table}[H]
    \scriptsize
    \caption{Processed GFS variables.}
    \label{tab:gfs_variables}
    \centering
    \begin{tabularx}{\textwidth}{l|l|l}
    \toprule
        \textbf{Description} & \textbf{Variable Name} & \textbf{Unit} \\ \midrule
        Precipitable water for entire atmosphere & \texttt{precipitable\_water\_entire\_atmosphere} & kg m$^{\text{-2}}$ \\
        Relative humidity 2m above ground & \texttt{relative\_humidity\_2m\_above\_ground} & \% \\
        Specific humidity 2m above ground & \texttt{specific\_humidity\_2m\_above\_ground} & Mass fraction \\
        Temperature 2m above ground & \texttt{temperature\_2m\_above\_ground} & $^\circ$C \\
        U component of wind 10m above ground & \texttt{u\_component\_of\_wind\_10m\_above\_ground} & m/s \\
        V component of wind 10m above ground & \texttt{v\_component\_of\_wind\_10m\_above\_ground} & m/s\\
        Downward shortwave radiation flux & \texttt{downward\_shortwave\_radiation\_flux} & W m$^{\text{-2}}$ \\
        Total cloud cover for entire atmosphere & \texttt{total\_cloud\_cover\_entire\_atmosphere} & \% \\
        Cumulative precipitation at surface for the previous 1--6 hours & \texttt{total\_precipitation\_surface} & kg m$^{\text{-2}}$ \\
        \bottomrule
    \end{tabularx}
\end{table}

\begin{table}[H]
\centering
\begin{threeparttable}
\caption{\textbf{Benchmark performance of the forecasting models at a 10-min lead time.} The best-performing value(s) for each site and evaluation metric are highlighted in \textbf{bold}. Ramp recall is evaluated using a 30\% relative error threshold.}
\label{tab:benchmark_10min}

\scriptsize
\setlength{\tabcolsep}{5pt}
\renewcommand{\arraystretch}{0.95}

\begin{tabularx}{\linewidth}{ll*{6}{>{\centering\arraybackslash}X}}
\toprule
\textbf{Site} &
\textbf{Model} &
\textbf{RMSE [W m$^{-2}$]} &
\textbf{FS [\%]} &
\textbf{rRMSE} &
\textbf{Ramp recall [\%]} &
\textbf{MAE [W m$^{-2}$]} &
\textbf{MBE [W m$^{-2}$]} \\
\midrule

\multirow{5}{*}{Akwatia} & Persistence & 126.5 & 0.0 & 0.291 & \textbf{8.7} & \textbf{69.2} & \textbf{0.7} \\
 & MLP & 121.2 & 4.2 & 0.278 & 5.0 & 77.7 & -14.3 \\
 & SUNSET & \textbf{120.1} & \textbf{5.1} & \textbf{0.276} & 0.6 & 82.2 & -11.2 \\
 & OmniVision & 130.8 & -3.4 & 0.3 & 5.0 & 79.0 & -24.0 \\
 & ViT & 127.6 & -0.8 & 0.293 & 5.0 & 80.6 & -15.1 \\
\addlinespace
\multirow{5}{*}{Golden} & Persistence & 107.5 & 0.0 & 0.209 & \textbf{7.1} & \textbf{45.1} & 2.7 \\
 & MLP & 108.9 & -1.3 & 0.212 & 5.6 & 63.8 & -3.6 \\
 & SUNSET & 91.7 & 14.7 & 0.178 & 6.3 & 45.9 & \textbf{-2.0} \\
 & OmniVision & 94.9 & 11.7 & 0.184 & 5.6 & 56.4 & 23.0 \\
 & ViT & \textbf{90.7} & \textbf{15.6} & \textbf{0.176} & 3.2 & 47.2 & 15.4 \\
\addlinespace
\multirow{5}{*}{Hong Kong} & Persistence & 120.6 & 0.0 & 0.288 & 2.7 & 60.7 & \textbf{-2.1} \\
 & MLP & 113.3 & 6.0 & 0.271 & \textbf{9.5} & 69.9 & -6.5 \\
 & SUNSET & 98.0 & 18.7 & 0.234 & 2.7 & 57.7 & -3.8 \\
 & OmniVision & 105.1 & 12.9 & 0.251 & 5.4 & 64.1 & -5.6 \\
 & ViT & \textbf{94.2} & \textbf{21.9} & \textbf{0.225} & 6.8 & \textbf{56.8} & -12.8 \\
\addlinespace
\multirow{5}{*}{Kologo} & Persistence & 75.1 & 0.0 & 0.194 & \textbf{7.8} & \textbf{38.4} & \textbf{-3.2} \\
 & MLP & 73.0 & 2.7 & 0.189 & 3.9 & 51.3 & -8.5 \\
 & SUNSET & 65.0 & 13.5 & 0.168 & 2.3 & 40.2 & -9.5 \\
 & OmniVision & 73.6 & 2.0 & 0.19 & 3.9 & 48.9 & -5.2 \\
 & ViT & \textbf{63.3} & \textbf{15.7} & \textbf{0.164} & 3.9 & 39.2 & -3.6 \\
\addlinespace
\multirow{5}{*}{Nottingham} & Persistence & 83.4 & 0.0 & 0.162 & 5.0 & \textbf{39.3} & 3.4 \\
 & MLP & 82.5 & 1.0 & 0.16 & 5.5 & 55.9 & -12.4 \\
 & SUNSET & 82.3 & 1.2 & 0.16 & 1.1 & 60.9 & -11.4 \\
 & OmniVision & 78.4 & 5.9 & 0.152 & \textbf{6.6} & 46.5 & \textbf{-0.0} \\
 & ViT & \textbf{69.3} & \textbf{16.9} & \textbf{0.134} & 1.1 & 43.2 & -6.8 \\
\addlinespace
\multirow{5}{*}{Paris} & Persistence & 105.4 & 0.0 & 0.231 & \textbf{7.1} & 46.7 & \textbf{-0.3} \\
 & MLP & 96.8 & 8.2 & 0.212 & 5.4 & 59.6 & -6.3 \\
 & SUNSET & 82.6 & 21.7 & \textbf{0.181} & 3.8 & 45.2 & -2.7 \\
 & OmniVision & 86.9 & 17.5 & 0.191 & 5.6 & \textbf{45.1} & -8.2 \\
 & ViT & \textbf{82.4} & \textbf{21.8} & \textbf{0.181} & 4.5 & 46.9 & -10.0 \\
\addlinespace
\multirow{5}{*}{Puy de Dôme} & Persistence & 102.3 & 0.0 & 0.222 & 5.4 & 44.9 & -2.2 \\
 & MLP & 100.2 & 2.0 & 0.218 & \textbf{6.2} & 58.9 & -7.3 \\
 & SUNSET & 83.5 & 18.4 & 0.182 & 3.5 & 44.5 & \textbf{-1.5} \\
 & OmniVision & 81.7 & 20.2 & 0.178 & 5.9 & 42.0 & -5.3 \\
 & ViT & \textbf{79.3} & \textbf{22.5} & \textbf{0.172} & 5.2 & \textbf{41.6} & -1.7 \\
\addlinespace
\multirow{5}{*}{Sankt Augustin} & Persistence & 109.8 & 0.0 & 0.267 & \textbf{6.1} & 53.1 & -3.0 \\
 & MLP & 99.2 & 9.7 & 0.241 & 4.0 & 61.2 & -6.0 \\
 & SUNSET & 92.4 & 15.8 & 0.225 & 3.9 & 52.3 & -2.9 \\
 & OmniVision & 91.6 & 16.5 & 0.223 & 5.4 & 52.1 & 8.8 \\
 & ViT & \textbf{90.5} & \textbf{17.6} & \textbf{0.22} & 3.2 & \textbf{50.2} & \textbf{2.0} \\

\bottomrule
\end{tabularx}

\begin{tablenotes}[flushleft]
\scriptsize
\item 
\end{tablenotes}
\end{threeparttable}
\end{table}

\begin{table}[H]
\centering
\begin{threeparttable}
\caption{\textbf{Benchmark performance of the forecasting models at a 30-min lead time.} The best-performing value(s) for each site and evaluation metric are highlighted in \textbf{bold}. Ramp recall is evaluated using a 30\% relative error threshold.}
\label{tab:benchmark_30min}

\scriptsize
\setlength{\tabcolsep}{5pt}
\renewcommand{\arraystretch}{0.95}

\begin{tabularx}{\linewidth}{ll*{6}{>{\centering\arraybackslash}X}}
\toprule
\textbf{Site} &
\textbf{Model} &
\textbf{RMSE [W m$^{-2}$]} &
\textbf{FS [\%]} &
\textbf{rRMSE} &
\textbf{Ramp recall [\%]} &
\textbf{MAE [W m$^{-2}$]} &
\textbf{MBE [W m$^{-2}$]} \\
\midrule

\multirow{5}{*}{Akwatia} & Persistence & 144.5 & 0.0 & 0.327 & \textbf{5.1} & \textbf{86.8} & \textbf{2.3} \\
 & MLP & 154.2 & -6.8 & 0.349 & 4.5 & 113.6 & -28.2 \\
 & SUNSET & 138.4 & 4.2 & 0.313 & 3.2 & 102.7 & -12.9 \\
 & OmniVision & 149.0 & -3.1 & 0.337 & 2.6 & 90.8 & -15.9 \\
 & ViT & \textbf{136.9} & \textbf{5.2} & \textbf{0.31} & 2.6 & 88.4 & -8.6 \\
\addlinespace
\multirow{5}{*}{Golden} & Persistence & 146.2 & 0.0 & 0.281 & \textbf{6.8} & 65.6 & 8.6 \\
 & MLP & 153.4 & -4.9 & 0.294 & 1.1 & 104.4 & -19.9 \\
 & SUNSET & 126.6 & 13.4 & 0.243 & 3.4 & \textbf{62.8} & \textbf{2.1} \\
 & OmniVision & 125.2 & 14.3 & 0.24 & 1.1 & 69.1 & 9.3 \\
 & ViT & \textbf{121.3} & \textbf{17.0} & \textbf{0.233} & 4.5 & 67.7 & -9.2 \\
\addlinespace
\multirow{5}{*}{Hong Kong} & Persistence & 133.1 & 0.0 & 0.32 & 11.5 & 71.8 & -3.5 \\
 & MLP & 137.7 & -3.5 & 0.331 & 7.7 & 96.1 & -22.4 \\
 & SUNSET & 112.4 & 15.6 & 0.27 & 1.9 & 70.8 & -8.0 \\
 & OmniVision & 120.1 & 9.7 & 0.289 & \textbf{13.5} & 77.4 & \textbf{-1.5} \\
 & ViT & \textbf{109.5} & \textbf{17.7} & \textbf{0.263} & 7.7 & \textbf{67.8} & -8.1 \\
\addlinespace
\multirow{5}{*}{Kologo} & Persistence & 82.8 & 0.0 & 0.211 & \textbf{12.1} & \textbf{46.5} & -6.5 \\
 & MLP & 101.0 & -22.0 & 0.257 & 1.6 & 79.2 & -15.5 \\
 & SUNSET & 77.9 & 5.9 & \textbf{0.198} & 0.0 & 48.2 & -13.6 \\
 & OmniVision & 81.0 & 2.2 & 0.206 & 8.9 & 54.6 & -25.5 \\
 & ViT & \textbf{77.6} & \textbf{6.2} & \textbf{0.198} & 0.8 & 47.0 & \textbf{-3.0} \\
\addlinespace
\multirow{5}{*}{Nottingham} & Persistence & 82.3 & 0.0 & 0.158 & 5.8 & \textbf{42.6} & 1.7 \\
 & MLP & 110.5 & -34.4 & 0.212 & 1.7 & 87.0 & -29.8 \\
 & SUNSET & 96.6 & -17.4 & 0.185 & 1.2 & 75.8 & -17.1 \\
 & OmniVision & 87.2 & -6.0 & 0.167 & \textbf{7.0} & 57.0 & \textbf{-0.5} \\
 & ViT & \textbf{77.2} & \textbf{6.1} & \textbf{0.148} & 0.0 & 48.9 & 3.3 \\
\addlinespace
\multirow{5}{*}{Paris} & Persistence & 110.0 & 0.0 & 0.239 & \textbf{7.0} & 54.4 & \textbf{1.7} \\
 & MLP & 109.6 & 0.4 & 0.238 & 4.7 & 77.1 & -10.1 \\
 & SUNSET & \textbf{91.2} & \textbf{17.1} & \textbf{0.198} & 3.2 & \textbf{53.3} & -4.2 \\
 & OmniVision & 95.3 & 13.4 & 0.207 & 4.3 & 57.0 & -5.7 \\
 & ViT & 92.6 & 15.8 & 0.201 & 3.0 & 57.9 & -13.2 \\
\addlinespace
\multirow{5}{*}{Puy de Dôme} & Persistence & 122.8 & 0.0 & 0.263 & \textbf{9.1} & 57.7 & -1.2 \\
 & MLP & 132.7 & -8.0 & 0.285 & 7.7 & 90.2 & -14.7 \\
 & SUNSET & 104.9 & 14.6 & 0.225 & 5.1 & 57.8 & \textbf{-0.9} \\
 & OmniVision & 106.0 & 13.7 & 0.227 & 7.2 & 57.8 & -2.3 \\
 & ViT & \textbf{102.4} & \textbf{16.6} & \textbf{0.22} & 3.4 & \textbf{56.3} & -2.8 \\
\addlinespace
\multirow{5}{*}{Sankt Augustin} & Persistence & 112.7 & 0.0 & 0.271 & \textbf{7.6} & 60.6 & \textbf{-3.2} \\
 & MLP & 116.9 & -3.7 & 0.281 & 4.4 & 83.6 & -11.0 \\
 & SUNSET & \textbf{99.2} & \textbf{12.0} & \textbf{0.239} & 3.7 & \textbf{60.2} & -6.4 \\
 & OmniVision & 99.4 & 11.9 & \textbf{0.239} & 7.4 & 61.7 & -6.5 \\
 & ViT & 99.9 & 11.4 & 0.24 & 2.5 & 63.1 & -3.4 \\

\bottomrule
\end{tabularx}

\begin{tablenotes}[flushleft]
\scriptsize
\item 
\end{tablenotes}
\end{threeparttable}
\end{table}

\begin{table}[H]
\centering
\begin{threeparttable}
\caption{\textbf{Benchmark performance of the forecasting models at a 60-min lead time.} The best-performing value(s) for each site and evaluation metric are highlighted in \textbf{bold}. Ramp recall is evaluated using a 30\% relative error threshold.}
\label{tab:benchmark_60min}

\scriptsize
\setlength{\tabcolsep}{5pt}
\renewcommand{\arraystretch}{0.95}

\begin{tabularx}{\linewidth}{ll*{6}{>{\centering\arraybackslash}X}}
\toprule
\textbf{Site} &
\textbf{Model} &
\textbf{RMSE [W m$^{-2}$]} &
\textbf{FS [\%]} &
\textbf{rRMSE} &
\textbf{Ramp recall [\%]} &
\textbf{MAE [W m$^{-2}$]} &
\textbf{MBE [W m$^{-2}$]} \\
\midrule

\multirow{5}{*}{Akwatia} & Persistence & 173.5 & 0.0 & 0.385 & \textbf{10.1} & \textbf{110.5} & \textbf{3.7} \\
 & MLP & 210.1 & -21.1 & 0.466 & 3.4 & 165.9 & -42.7 \\
 & SUNSET & 166.9 & 3.8 & 0.37 & 0.7 & 132.5 & -10.7 \\
 & OmniVision & 168.1 & 3.1 & 0.373 & 9.4 & 121.4 & -7.8 \\
 & ViT & \textbf{156.7} & \textbf{9.7} & \textbf{0.347} & 1.3 & 112.1 & -22.2 \\
\addlinespace
\multirow{5}{*}{Golden} & Persistence & 172.4 & 0.0 & 0.326 & 1.7 & 83.6 & 17.2 \\
 & MLP & 212.6 & -23.4 & 0.402 & 0.0 & 168.8 & -33.7 \\
 & SUNSET & 146.4 & 15.1 & 0.277 & 3.4 & \textbf{75.2} & \textbf{-2.9} \\
 & OmniVision & 150.5 & 12.7 & 0.284 & \textbf{5.1} & 89.6 & 16.0 \\
 & ViT & \textbf{138.2} & \textbf{19.8} & \textbf{0.261} & 0.0 & 75.5 & -4.3 \\
\addlinespace
\multirow{5}{*}{Hong Kong} & Persistence & 149.8 & 0.0 & 0.335 & 2.4 & 87.8 & \textbf{-5.2} \\
 & MLP & 180.4 & -20.4 & 0.404 & 0.0 & 138.7 & -40.9 \\
 & SUNSET & \textbf{123.3} & \textbf{17.7} & \textbf{0.276} & 4.8 & \textbf{85.4} & -10.1 \\
 & OmniVision & 139.3 & 7.0 & 0.312 & \textbf{11.9} & 99.3 & -20.1 \\
 & ViT & 126.3 & 15.7 & 0.283 & 9.5 & 87.0 & 7.4 \\
\addlinespace
\multirow{5}{*}{Kologo} & Persistence & 101.9 & 0.0 & 0.254 & \textbf{6.0} & \textbf{58.7} & \textbf{-11.6} \\
 & MLP & 144.2 & -41.5 & 0.36 & 5.2 & 118.2 & -32.2 \\
 & SUNSET & \textbf{92.9} & \textbf{8.9} & \textbf{0.232} & 0.9 & 59.8 & -24.3 \\
 & OmniVision & 104.4 & -2.4 & 0.26 & 3.4 & 76.8 & -47.6 \\
 & ViT & 96.3 & 5.5 & 0.24 & 0.9 & 59.3 & -21.1 \\
\addlinespace
\multirow{5}{*}{Nottingham} & Persistence & 101.4 & 0.0 & 0.192 & 6.0 & 52.3 & \textbf{3.7} \\
 & MLP & 148.1 & -46.1 & 0.28 & 2.7 & 122.9 & -14.1 \\
 & SUNSET & 111.5 & -10.0 & 0.211 & 0.0 & 90.3 & -20.0 \\
 & OmniVision & 125.7 & -24.1 & 0.238 & \textbf{6.7} & 91.2 & -25.3 \\
 & ViT & \textbf{86.9} & \textbf{14.3} & \textbf{0.164} & 0.0 & \textbf{52.2} & 13.1 \\
\addlinespace
\multirow{5}{*}{Paris} & Persistence & 121.6 & 0.0 & 0.259 & \textbf{6.7} & 65.2 & 6.4 \\
 & MLP & 142.1 & -16.9 & 0.303 & 4.1 & 112.0 & -21.8 \\
 & SUNSET & \textbf{101.1} & \textbf{16.8} & \textbf{0.216} & 2.2 & 63.7 & -8.4 \\
 & OmniVision & 104.9 & 13.8 & 0.223 & 5.4 & \textbf{63.5} & \textbf{-1.3} \\
 & ViT & 103.2 & 15.1 & 0.22 & 2.1 & 67.5 & -10.4 \\
\addlinespace
\multirow{5}{*}{Puy de Dôme} & Persistence & 139.6 & 0.0 & 0.294 & \textbf{6.2} & \textbf{70.5} & \textbf{1.8} \\
 & MLP & 169.6 & -21.5 & 0.357 & 5.4 & 127.7 & -29.9 \\
 & SUNSET & \textbf{120.0} & \textbf{14.1} & \textbf{0.252} & 2.3 & 70.7 & -8.2 \\
 & OmniVision & 125.0 & 10.5 & 0.263 & 5.0 & 72.5 & -4.8 \\
 & ViT & 120.9 & 13.4 & 0.254 & 1.8 & 73.2 & -10.4 \\
\addlinespace
\multirow{5}{*}{Sankt Augustin} & Persistence & 121.5 & 0.0 & 0.288 & 6.6 & 70.0 & -2.5 \\
 & MLP & 136.7 & -12.5 & 0.324 & 3.5 & 106.6 & \textbf{1.0} \\
 & SUNSET & 108.5 & 10.7 & 0.257 & 3.6 & 70.2 & -3.7 \\
 & OmniVision & 121.0 & 0.4 & 0.287 & \textbf{6.8} & 84.6 & -18.6 \\
 & ViT & \textbf{105.2} & \textbf{13.5} & \textbf{0.249} & 1.4 & \textbf{68.6} & -13.3 \\

\bottomrule
\end{tabularx}

\begin{tablenotes}[flushleft]
\scriptsize
\item 
\end{tablenotes}
\end{threeparttable}
\end{table}

\begin{table}[H]
\centering
\begin{threeparttable}
\caption{\textbf{Benchmark performance of the forecasting models at a 90-min lead time.} The best-performing value(s) for each site and evaluation metric are highlighted in \textbf{bold}. Ramp recall is evaluated using a 30\% relative error threshold.}
\label{tab:benchmark_90min}

\scriptsize
\setlength{\tabcolsep}{5pt}
\renewcommand{\arraystretch}{0.95}

\begin{tabularx}{\linewidth}{ll*{6}{>{\centering\arraybackslash}X}}
\toprule
\textbf{Site} &
\textbf{Model} &
\textbf{RMSE [W m$^{-2}$]} &
\textbf{FS [\%]} &
\textbf{rRMSE} &
\textbf{Ramp recall [\%]} &
\textbf{MAE [W m$^{-2}$]} &
\textbf{MBE [W m$^{-2}$]} \\
\midrule

\multirow{5}{*}{Akwatia} & Persistence & \textbf{175.0} & \textbf{0.0} & \textbf{0.38} & \textbf{5.5} & \textbf{118.8} & \textbf{3.3} \\
 & MLP & 245.8 & -40.4 & 0.534 & 0.7 & 196.9 & -53.8 \\
 & SUNSET & 185.2 & -5.8 & 0.403 & 2.1 & 150.9 & -15.9 \\
 & OmniVision & 208.6 & -19.2 & 0.454 & 3.4 & 161.0 & -40.1 \\
 & ViT & 182.7 & -4.4 & 0.397 & 2.1 & 136.4 & -3.4 \\
\addlinespace
\multirow{5}{*}{Golden} & Persistence & 194.2 & 0.0 & 0.358 & 1.7 & 99.7 & 27.9 \\
 & MLP & 244.3 & -25.8 & 0.451 & 0.0 & 194.9 & -40.6 \\
 & SUNSET & 159.2 & 18.0 & 0.294 & 0.0 & \textbf{85.1} & 4.3 \\
 & OmniVision & \textbf{153.2} & \textbf{21.1} & \textbf{0.283} & \textbf{5.0} & 92.7 & \textbf{-2.7} \\
 & ViT & 161.5 & 16.9 & 0.298 & 1.7 & 88.3 & 8.4 \\
\addlinespace
\multirow{5}{*}{Hong Kong} & Persistence & 170.5 & 0.0 & 0.382 & \textbf{19.2} & 100.6 & -7.8 \\
 & MLP & 225.6 & -32.3 & 0.505 & 11.5 & 173.0 & -53.1 \\
 & SUNSET & \textbf{136.9} & \textbf{19.7} & \textbf{0.307} & 0.0 & 100.1 & \textbf{-5.1} \\
 & OmniVision & 149.8 & 12.2 & 0.335 & 3.8 & 112.7 & -12.7 \\
 & ViT & 137.2 & 19.6 & \textbf{0.307} & 7.7 & \textbf{100.0} & 15.5 \\
\addlinespace
\multirow{5}{*}{Kologo} & Persistence & 113.0 & 0.0 & 0.277 & 5.3 & 65.1 & -16.7 \\
 & MLP & 184.0 & -62.8 & 0.451 & \textbf{6.1} & 154.4 & -51.1 \\
 & SUNSET & 105.7 & 6.4 & 0.259 & 0.9 & 69.2 & -36.3 \\
 & OmniVision & 114.5 & -1.4 & 0.281 & 5.3 & 81.4 & -26.5 \\
 & ViT & \textbf{104.4} & \textbf{7.7} & \textbf{0.256} & 0.0 & \textbf{63.1} & \textbf{-4.8} \\
\addlinespace
\multirow{5}{*}{Nottingham} & Persistence & 109.9 & 0.0 & 0.205 & 3.6 & \textbf{59.2} & \textbf{1.9} \\
 & MLP & 187.3 & -70.3 & 0.349 & 0.7 & 160.0 & -32.2 \\
 & SUNSET & 119.4 & -8.6 & 0.223 & 0.0 & 98.0 & -23.6 \\
 & OmniVision & 115.0 & -4.6 & 0.215 & \textbf{6.4} & 80.2 & -24.0 \\
 & ViT & \textbf{92.1} & \textbf{16.3} & \textbf{0.172} & 0.0 & 61.0 & -19.7 \\
\addlinespace
\multirow{5}{*}{Paris} & Persistence & 129.0 & 0.0 & 0.27 & \textbf{7.1} & 72.6 & 11.6 \\
 & MLP & 178.5 & -38.4 & 0.374 & 5.2 & 146.1 & -33.4 \\
 & SUNSET & \textbf{106.9} & \textbf{17.1} & \textbf{0.224} & 3.4 & \textbf{70.1} & -15.0 \\
 & OmniVision & 110.9 & 14.0 & 0.232 & 5.3 & 70.3 & \textbf{-7.8} \\
 & ViT & 108.4 & 16.0 & 0.227 & 1.8 & 72.2 & -9.7 \\
\addlinespace
\multirow{5}{*}{Puy de Dôme} & Persistence & 147.3 & 0.0 & 0.305 & 5.2 & \textbf{78.7} & \textbf{6.2} \\
 & MLP & 206.0 & -39.8 & 0.426 & 4.8 & 164.2 & -44.2 \\
 & SUNSET & 130.0 & 11.7 & 0.269 & 1.8 & 81.8 & -12.2 \\
 & OmniVision & 133.1 & 9.7 & 0.275 & \textbf{5.8} & 82.3 & 10.6 \\
 & ViT & \textbf{129.6} & \textbf{12.0} & \textbf{0.268} & 3.8 & 78.8 & -10.1 \\
\addlinespace
\multirow{5}{*}{Sankt Augustin} & Persistence & 132.0 & 0.0 & 0.308 & \textbf{5.7} & 79.1 & -1.5 \\
 & MLP & 159.9 & -21.1 & 0.373 & 4.5 & 125.5 & -14.6 \\
 & SUNSET & 118.3 & 10.4 & 0.276 & 3.7 & 79.8 & -7.7 \\
 & OmniVision & 121.3 & 8.2 & 0.283 & \textbf{5.7} & 82.9 & -5.2 \\
 & ViT & \textbf{113.6} & \textbf{14.0} & \textbf{0.265} & 1.4 & \textbf{73.4} & \textbf{0.7} \\

\bottomrule
\end{tabularx}

\begin{tablenotes}[flushleft]
\scriptsize
\item 
\end{tablenotes}
\end{threeparttable}
\end{table}

\begin{table}[H]
\centering
\begin{threeparttable}
\caption{\textbf{Benchmark performance of the forecasting models at a 120-min lead time.} The best-performing value(s) for each site and evaluation metric are highlighted in \textbf{bold}. Ramp recall is evaluated using a 30\% relative error threshold.}
\label{tab:benchmark_120min}

\scriptsize
\setlength{\tabcolsep}{5pt}
\renewcommand{\arraystretch}{0.95}

\begin{tabularx}{\linewidth}{ll*{6}{>{\centering\arraybackslash}X}}
\toprule
\textbf{Site} &
\textbf{Model} &
\textbf{RMSE [W m$^{-2}$]} &
\textbf{FS [\%]} &
\textbf{rRMSE} &
\textbf{Ramp recall [\%]} &
\textbf{MAE [W m$^{-2}$]} &
\textbf{MBE [W m$^{-2}$]} \\
\midrule

\multirow{5}{*}{Akwatia} & Persistence & \textbf{191.8} & \textbf{0.0} & \textbf{0.411} & \textbf{6.2} & \textbf{137.1} & 3.3 \\
 & MLP & 288.3 & -50.3 & 0.617 & 1.4 & 233.8 & -67.4 \\
 & SUNSET & 201.5 & -5.0 & 0.431 & 0.7 & 160.5 & -6.4 \\
 & OmniVision & 205.7 & -7.2 & 0.44 & 4.2 & 154.7 & \textbf{-1.1} \\
 & ViT & 199.3 & -3.9 & 0.427 & 3.5 & 147.9 & 15.3 \\
\addlinespace
\multirow{5}{*}{Golden} & Persistence & 205.4 & 0.0 & 0.372 & 0.0 & 110.8 & 38.9 \\
 & MLP & 277.3 & -35.0 & 0.502 & \textbf{4.2} & 229.1 & -63.5 \\
 & SUNSET & 165.6 & 19.4 & 0.3 & 0.0 & 93.6 & \textbf{2.4} \\
 & OmniVision & \textbf{159.7} & \textbf{22.3} & \textbf{0.289} & 0.0 & 95.0 & 11.6 \\
 & ViT & 161.5 & 21.4 & 0.292 & 0.0 & \textbf{91.0} & -6.5 \\
\addlinespace
\multirow{5}{*}{Hong Kong} & Persistence & 184.2 & 0.0 & 0.393 & 3.1 & 111.7 & \textbf{-0.8} \\
 & MLP & 263.8 & -43.2 & 0.562 & 6.2 & 207.0 & -70.6 \\
 & SUNSET & 152.4 & 17.3 & 0.325 & 0.0 & 117.3 & -3.2 \\
 & OmniVision & 152.7 & 17.1 & 0.326 & \textbf{9.4} & 112.4 & -1.9 \\
 & ViT & \textbf{147.6} & \textbf{19.9} & \textbf{0.315} & \textbf{9.4} & \textbf{107.7} & 8.6 \\
\addlinespace
\multirow{5}{*}{Kologo} & Persistence & 132.5 & 0.0 & 0.319 & \textbf{5.4} & 78.9 & \textbf{-22.4} \\
 & MLP & 232.0 & -75.1 & 0.559 & 0.9 & 194.6 & -85.4 \\
 & SUNSET & 116.1 & 12.4 & 0.28 & 1.8 & \textbf{75.3} & -43.4 \\
 & OmniVision & \textbf{114.3} & \textbf{13.7} & \textbf{0.275} & 0.9 & 77.7 & -40.8 \\
 & ViT & 123.3 & 7.0 & 0.297 & 1.8 & 80.6 & -53.8 \\
\addlinespace
\multirow{5}{*}{Nottingham} & Persistence & 110.3 & 0.0 & 0.205 & \textbf{6.5} & 61.8 & \textbf{0.2} \\
 & MLP & 223.7 & -102.9 & 0.416 & 5.6 & 191.4 & -59.3 \\
 & SUNSET & 136.6 & -23.9 & 0.254 & 0.0 & 117.0 & -18.9 \\
 & OmniVision & 110.1 & 0.2 & 0.205 & \textbf{6.5} & 78.9 & -20.9 \\
 & ViT & \textbf{86.3} & \textbf{21.7} & \textbf{0.161} & 2.4 & \textbf{58.8} & -3.7 \\
\addlinespace
\multirow{5}{*}{Paris} & Persistence & 136.5 & 0.0 & 0.281 & \textbf{5.8} & 79.1 & 17.8 \\
 & MLP & 210.4 & -54.2 & 0.433 & 5.5 & 174.6 & -56.3 \\
 & SUNSET & 113.3 & 17.0 & 0.233 & 2.8 & 76.2 & -6.7 \\
 & OmniVision & 118.1 & 13.4 & 0.243 & 2.8 & 79.7 & -6.1 \\
 & ViT & \textbf{112.0} & \textbf{17.9} & \textbf{0.231} & 1.8 & \textbf{75.6} & \textbf{-2.5} \\
\addlinespace
\multirow{5}{*}{Puy de Dôme} & Persistence & 164.0 & 0.0 & 0.334 & 5.8 & \textbf{90.1} & 11.7 \\
 & MLP & 259.1 & -57.9 & 0.527 & 2.8 & 211.6 & -76.3 \\
 & SUNSET & \textbf{140.3} & \textbf{14.4} & \textbf{0.286} & 3.1 & 92.0 & -17.8 \\
 & OmniVision & 144.3 & 12.0 & 0.294 & \textbf{6.8} & 90.6 & \textbf{-10.3} \\
 & ViT & 141.9 & 13.5 & 0.289 & 2.1 & 92.4 & -22.2 \\
\addlinespace
\multirow{5}{*}{Sankt Augustin} & Persistence & 135.0 & 0.0 & 0.311 & \textbf{7.8} & 82.0 & \textbf{0.1} \\
 & MLP & 181.8 & -34.7 & 0.42 & 4.6 & 144.6 & -20.4 \\
 & SUNSET & 124.7 & 7.6 & 0.288 & 3.2 & 85.5 & -6.9 \\
 & OmniVision & 126.4 & 6.4 & 0.292 & 6.9 & 92.1 & -14.2 \\
 & ViT & \textbf{119.3} & \textbf{11.6} & \textbf{0.275} & 2.7 & \textbf{81.0} & -7.0 \\

\bottomrule
\end{tabularx}

\begin{tablenotes}[flushleft]
\scriptsize
\item 
\end{tablenotes}
\end{threeparttable}
\end{table}

\bibliography{bibliography}